%% file: main.tex
\documentclass[10pt,twocolumn,letterpaper]{article}
\usepackage[pagenumbers, final]{cvpr}      % To produce the FINAL version

\usepackage{graphicx}
\usepackage{amsmath}
\usepackage{amssymb}
\usepackage{booktabs}
\usepackage{currfile}

\usepackage[numbers]{natbib}  % only use \citet or \citep, never \cite
\usepackage{enumitem}  % no-indent itemize environments
\usepackage{xcolor}  % colors, duh
\usepackage{color}
\usepackage{colortbl}
\usepackage{microtype}
\usepackage{nicefrac}
\usepackage{siunitx}
\usepackage[hang,flushmargin]{footmisc}

\usepackage[breaklinks,colorlinks]{hyperref}

\usepackage[capitalize]{cleveref}
\crefname{section}{Sec.}{Secs.}
\Crefname{section}{Section}{Sections}
\Crefname{table}{Table}{Tables}
\crefname{table}{Tab.}{Tabs.}

\def\cvprPaperID{804}
\def\confName{CVPR}
\def\confYear{2022}

\makeatletter
\newcommand{\printfnsymbol}[1]{%
  \textsuperscript{\@fnsymbol{#1}}%
}
\makeatother

\input{preamble.tex}
\begin{document}

% Nicer margins for equations.
\abovedisplayskip=6pt
\abovedisplayshortskip=6pt
\belowdisplayskip=6pt
\belowdisplayshortskip=6pt

\title{\litlong{}: \\ Geometry-Free Novel View Synthesis Through Set-Latent Scene Representations}

\renewcommand*{\thefootnote}{\fnsymbol{footnote}}

\author{
Mehdi S. M. Sajjadi
\and
Henning Meyer
\and
Etienne Pot
\and
Urs Bergmann
\and
Klaus Greff
\and
Noha Radwan
\and
Suhani Vora
\and
Mario Lu\v{c}i\'c
\and
Daniel Duckworth
\and
Alexey Dosovitskiy\footnotemark[1]
\and
Jakob Uszkoreit\footnotemark[1]
\and
Thomas Funkhouser
\and
Andrea Tagliasacchi\smash{\footnotemark[3]\;\footnotemark[1]}
}
\affiliation{Google Research\qquad \footnotemark[3]\; Simon Fraser University}

\maketitle

\footnotetext[1]{Work done while at Google.}
\footnotetext{
Contributions:
\emph{MS}: original idea \& conceptualization, main implementation, experiments, writing, organization, lead;
\emph{HM}: idea conceptualization, implementation, experiments, writing;
\emph{EP}: implementation, experiments, writing;
\emph{UB}: idea conceptualization, implementation, experiments, writing;
\emph{KG}: dataset support, advising;
\emph{NR}: project-independent code;
\emph{SV}: project-independent code;
\emph{ML}: advising, writing;
\emph{DD}: advising, project-independent code;
\emph{AD}: advising;
\emph{JU}: advising;
\emph{TF}: datasets, advising, writing;
\emph{AT}: advising, writing.
\newline
Correspondence:
\href{mailto:srt@msajjadi.com}{srt@msajjadi.com} \& \href{mailto:tutmann@google.com}{tutmann@google.com}
\newline
Project website:
\litwebsite{}
}

\renewcommand{\cite}{\citep}
\renewcommand{\citet}{\citep}

\input{tex/0-abstract}
\input{tex/1-intro}
\input{tex/2-related}
\input{tex/3-method}

\input{tex/4-exps}
\input{tex/5-conclusion}

\input{tex/6-acknowledgements}

{
    \small
    \bibliographystyle{plainnat}
    \bibliography{main}
}

\clearpage
\input{tex/8-appendix}

\end{document}

%% file: preamble.tex
\def \litlong {Scene Representation Transformer}
\def \lit {SRT}
\def \uplit {Up\lit{}}
\def \vlit {V-\lit{}}
\def \litwebsite {\href{https://srt-paper.github.io/}{srt-paper.github.io}}

\def \gsr {scene representation}
\def \msn {MSN}

\def \patchsize {K}

\def \numoctaves {L}  % following nerf

\def \rayo {o}
\def \rayd {d}

\def \posenc {\gamma}  % following nerf

\newcommand{\rot}[2][l]{\rotatebox[origin=#1]{90}{#2}}  % rotate stuff 90 degrees

\definecolor{turquoise}{cmyk}{0.65,0,0.1,0.3}
\definecolor{purple}{rgb}{0.65,0,0.65}
\definecolor{dark_green}{rgb}{0, 0.5, 0}
\definecolor{orange}{rgb}{0.8, 0.6, 0.2}
\definecolor{dark_orange}{rgb}{0.7, 0.6, 0.3}
\definecolor{red}{rgb}{0.8, 0.2, 0.2}
\definecolor{darkred}{rgb}{0.6, 0.1, 0.05}
\definecolor{blueish}{rgb}{0.0, 0.3, .6}
\definecolor{light_gray}{rgb}{0.7, 0.7, .7}
\definecolor{pink}{rgb}{1, 0, 1}
\definecolor{cyan}{rgb}{0., 1, 1}

\definecolor{checkyes}{rgb}{0.7, 1.0, 0.7}
\definecolor{crossno}{rgb}{1.0, 0.7, 0.7}
\def \y {$\checkmark$\cellcolor{checkyes}}
\def \n {$\times$\cellcolor{crossno}}

\definecolor{tabbestcolor}{rgb}{0.785, 0.851, 0.969}
\def \best {\cellcolor{tabbestcolor!85}}
\def \sbest {\cellcolor{tabbestcolor!30}}

\definecolor{mydarkblue}{rgb}{0,0.08,0.55}
\hypersetup{
    colorlinks=true,
    linkcolor=mydarkblue,
    citecolor=mydarkblue,
    filecolor=mydarkblue,
    urlcolor=mydarkblue
}

\DeclareMathOperator*{\argmin}{arg\,min}

\newcommand{\expect}{\mathbb{E}}
\newcommand{\real}{\mathbb{R}}

\renewcommand{\paragraph}[1]{\vspace{.2em}\noindent\textbf{#1}.}

\newcommand{\scene}{\{\code_\iCode\}}
\newcommand{\iScene}{s}
\newcommand{\image}{\mathbf{I}}

\newcommand{\iImage}{i}
\newcommand{\nImages}{I}
\newcommand{\code}{\mathbf{z}}
\newcommand{\iCode}{z}
\newcommand{\nCodes}{Z}
\newcommand{\dCode}{d}
\newcommand{\encoder}{\mathcal{E}}
\newcommand{\pars}{\theta}
\newcommand{\given}{~|~}
\newcommand{\decoder}{\mathcal{D}}
\newcommand{\camera}{\mathbf{c}}
\newcommand{\ray}{\mathbf{r}}
\newcommand{\rayOrigin}{\mathbf{o}}
\newcommand{\rayDir}{\mathbf{d}}
\newcommand{\C}{\mathbf{c}}
\newcommand{\concat}{\oplus}
\newcommand{\W}{W}
\renewcommand{\H}{H}
\renewcommand{\C}{C}
\newcommand{\h}{h}
\newcommand{\w}{w}
\renewcommand{\c}{\textsc{c}}
\renewcommand{\C}{C}
\newcommand{\Feature}{\mathbf{F}}
\newcommand{\Featurehat}{\mathbf{\hat F}}
\newcommand{\patchFeature}{\mathbf{f}}
\newcommand{\iPatches}{f}

\newcommand{\backbone}{\text{CNN}}
\newcommand{\bias}{\mathcal{B}}

\newcommand{\patchind}{n}

%% file: tex/0-abstract.tex
\begin{abstract}
A classical problem in computer vision is to infer a 3D scene representation from few images that can be used to render novel views at interactive rates.
Previous work focuses on reconstructing pre-defined 3D representations, \eg textured meshes, or implicit representations, \eg radiance fields, and often requires input images with precise camera poses and long processing times for each novel scene.

In this work, we propose the \litlong{} (\lit{}), a method which processes posed or unposed RGB images of a new area, infers a ``set-latent scene representation'', and synthesises novel views, all in a single feed-forward pass.
To calculate the scene representation, we propose a generalization of the Vision Transformer to sets of images, enabling global information integration, and hence 3D reasoning.
An efficient decoder transformer parameterizes the light field by attending into the scene representation to render novel views.
Learning is supervised end-to-end by minimizing a novel-view reconstruction error.

We show that this method outperforms recent baselines in terms of PSNR and speed on synthetic datasets, including a new dataset created for the paper.
Further, we demonstrate that \lit{} scales to support interactive visualization and semantic segmentation of real-world outdoor environments using  Street View imagery.

% \vspace{-5mm}
\end{abstract}

%% file: tex/1-intro.tex
\section{Introduction}
\label{sec:introduction}
\vspace{-2mm}

The goal of our work is interactive novel view synthesis: given \emph{few} RGB images of a previously unseen scene, we synthesize novel views of the same scene at interactive rates and without expensive per-scene processing.
Such a system can be useful for virtual exploration of urban spaces~\cite{mirowski2019streetlearn}, as well as other mapping, visualization, and AR/VR applications~\cite{musialski2013survey}.
The main challenge here is to infer a scene representation that encodes enough 3D information to render novel views with correct parallax and occlusions.

\input{figs/teaser}

Traditional methods build explicit 3D representations, such as colored point clouds~\cite{agarwal2011building}, meshes~\cite{romanoni2017mesh}, voxels~\cite{shim20113d}, octrees~\cite{truong2014octree}, and multi-plane images~\cite{Zhou18siggraph_mpi}.
These representations enable interactive rendering, but they usually require expensive and fragile reconstruction processes.

More recent work has investigated representing scenes with purely
implicit representations~\cite{Tewari20egsr_neural_rendering}.
Notably, NeRF represents the scene as a 3D volume parameterized by an MLP~\cite{Mildenhall20eccv_nerf}, and has been demonstrated to scale to challenging real-world settings~\cite{MartinBrualla21cvpr_nerfw}.
However, it typically requires hundreds of MLP evaluations for the volumetric rendering of each ray, relies on accurate camera poses, and requires an expensive training procedure to onboard new scenes, as no parameters are shared across different scenes.

Follow-ups that address these shortcomings include re-projection based methods which still rely on accurate camera poses due to their explicit use of geometry, and latent models which are usually geometry-free and able to reason globally, an advantage for the sparse input view setting.
However, those methods generally fail to scale to complex real-world datasets.
As shown in \cref{tab:baselines}, interactive novel view synthesis on complex real-world data remains a challenge.

We tackle this challenge by employing an encoder-decoder model built on transformers, learning a scalable implicit representation, and replacing explicit geometric operations with \emph{learned} attention mechanisms.
Different from a number of prior works that learn the latent scene representations through an auto-decoder optimization procedure~\citep{sitzmann2021lfn, Jang21iccv_CodeNeRF}, we specifically rely on an encoder architecture to allow for \emph{instant} inference on novel scenes.
At the same time, we do not rely on explicit locally-conditioned geometry~\citep{Yu21cvpr_pixelNeRF, Trevithick21iccv_GRF} to instead allow the model to reason globally.
This not only comes with the advantage of stronger generalization abilities (\eg, for rendering novel cameras further away from the input views, see \cref{fig:multishapenet}), but also enables the model to be more efficient, as global information is processed once per scene (\eg, to reconstruct 3D geometry and to resolve occlusions) in our model, rather than once or hundreds of times per rendered pixel, as in previous methods~\cite{Yu21cvpr_pixelNeRF, Trevithick21iccv_GRF, Wang21cvpr_IBRNet}.

We evaluate the proposed model on several datasets of increasing complexity against relevant prior art, and see that it strikes a unique balance between scalability to complex scenes (\cref{sec:exp}), robustness to noisy camera poses (or no poses at all, \cref{plot:camnoise}), and efficiency in interactive applications (\cref{tab:benchmark}).
Further, we show a proof-of-concept for the downstream task of semantic segmentation using the learned representation on a challenging real-world dataset (\cref{fig:sv}).

%% file: figs/teaser.tex
\begin{figure}
\centering
\includegraphics[clip,trim=48mm 123mm 75mm 82mm, width=\linewidth]{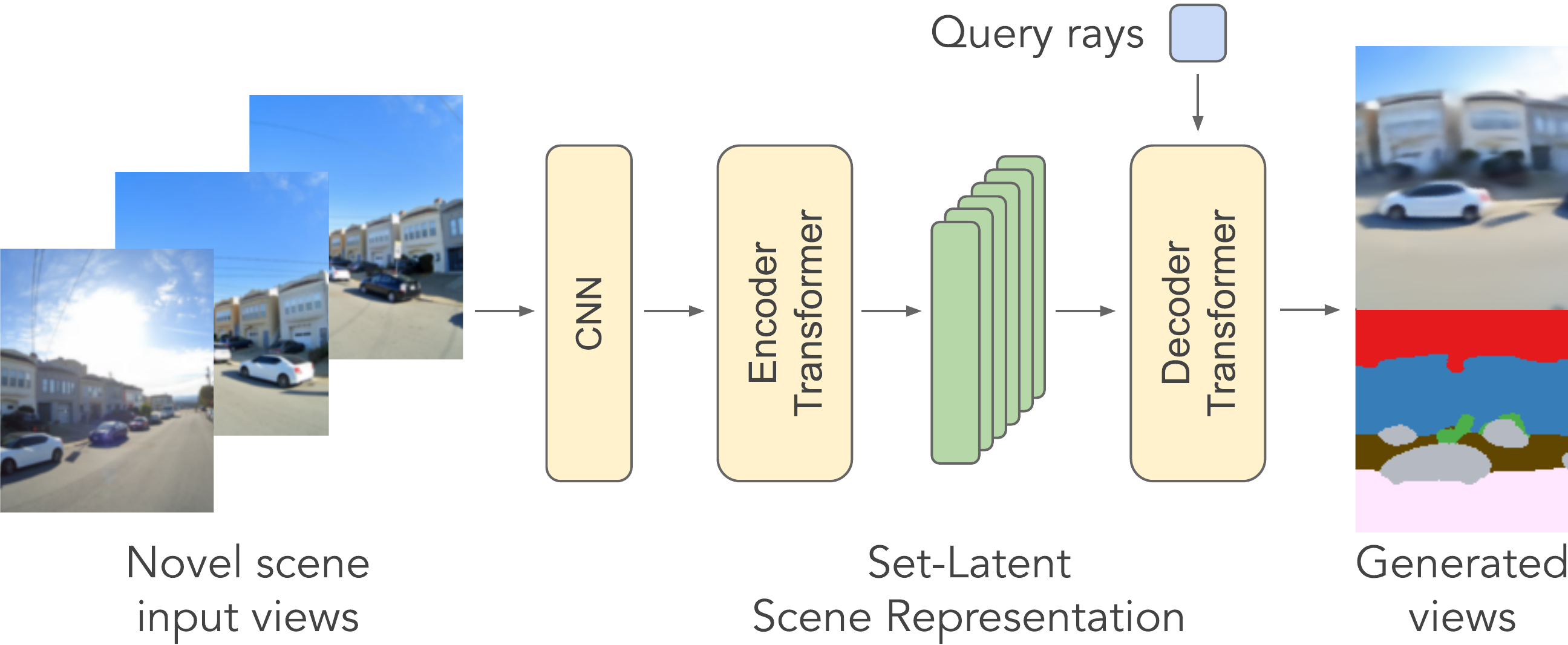}
\caption{
    \textbf{Model overview} --
    \lit{} encodes a collection of images into the scene representation: a \textit{set} of latent features.
    Novel views are rendered in real-time by attending into the latent representation with light field rays, see~\cref{fig:architecture} for details.
\vspace{-5mm}
}
\label{fig:teaser}
\end{figure}

%% file: tex/2-related.tex
\input{tables/baselines}
\section{Related Work}
\label{sec:relatedworks}

There is a long history of prior work on novel view synthesis from multiple RGB images based on neural representations~(see~\cref{tab:baselines} and~\cite{tewari2021arxiv_survey}).
Recent work has demonstrated that novel views can be synthesized using deep networks trained to compute radiance values for a given 3D position and direction.
In particular, NeRF~\citep{Mildenhall20eccv_nerf} trains an MLP to store radiance and density used in a volume rendering system and produces novel views with remarkably fine details.
Several extensions have been proposed to mitigate some its shortcomings, including ones to accelerate rendering~\cite{Hedman21iccv_baking_NeRF,  Reiser21iccv_KiloNeRF, Yu21iccv_PlenOctrees}, handle color variations~\cite{MartinBrualla21cvpr_nerfw}, optimize camera poses~\cite{Lin21iccv_BARF}, and perform well with few images~\citep{Jain21iccv_DietNeRF}.
However, these methods generally still require an expensive optimization of an MLP for \emph{each} novel scene, and thus are impractical for deployment at large scale.

\paragraph{NeRF with Re-projection}
Several NeRF extensions have been proposed that do not always require to be trained individually for each novel scene.
These methods rely on explicit 3D-to-2D projections of samples along the rays to gather features extracted from CNNs, which are then aggregated by a neural network for each point~\citep{Trevithick21iccv_GRF, Wang21cvpr_IBRNet, Yu21cvpr_pixelNeRF}, or into a voxel grid~\citep{Chen21iccv_MVSNeRF}.
As a result, these networks can be pre-trained, and then applied to novel scenes either without any optimization, or with additional per-scene fine-tuning.
However, reliance on explicit camera projections requires precise camera poses, and direct projections cannot take occlusions of objects between samples and input features into account.

\paragraph{Light Fields}
Other methods learn to represent a scene with a latent vector from which novel views can be synthesized~\citep{eslami2018neural}.
More recently, LFN~\citep{sitzmann2021lfn} use an auto-decoder to infer a latent code from input images, which then conditions an MLP for computing radiance values for rays of novel views.
This method is several times faster than volumetric rendering due to the fact that the network is only queried \emph{once} per pixel, rather than hundreds of times for volumetric rendering.
However, it does not scale beyond simple controlled settings, requires an expensive optimization to produce a latent code for a novel scene, and therefore also relies upon accurate input camera poses.
In contrast, we utilize an encoder to infer the scene representation, which executes nearly instantly in comparison (see \cref{tab:benchmark}), and can leverage, but does not require, camera poses (see~\cref{sec:exp:unposed}).

\input{figs/architecture}

\paragraph{Transformers}
Originally proposed for seq-to-seq tasks in natural language processing, transformers~\citep{transformer} drop the recurrent memory of prior methods, and instead use a multi-head attention mechanism to gather information from the most relevant portions of the input.
The vision community has deeply researched applications thereof in image and video processing, where they have meanwhile achieved state of the art in several tasks~\citep{transformersurvey}, including inherently geometric applications such as depth estimation~\citep{vitdepth}, point cloud processing~\cite{pointtransformer}, and generative geometry-free view synthesis~\citep{gfvs}.
The latter work uses transformers to learn a sequence of VQ-GAN embeddings~\citep{vqgan} to sample realistic views given a single image.
While results are of high-quality, this method is slow due to its auto-regressive nature, and cannot be used for video rendering, as each frame is sampled independently, creating strong flickering artifacts.
ViT~\citep{vit} has been proposed for very large-scale classification tasks: local features are extracted from patches of the input image, and fed to a transformer network.
We will demonstrate that this architecture is well-suited for learning 3D scene representations by building on top of this idea, extending it to ingest patches from multiple images, and generalizing it to 3D by adding a transformer decoder queried with rays.

%% file: tables/baselines.tex
\begin{table}[t]
\centering
\setlength{\tabcolsep}{3pt}
\def\arraystretch{1}
\resizebox{\linewidth}{!}{
\begin{tabular}{lccccc|cccc|ccc|cc|c}
\toprule
 & \multicolumn{5}{c}{NeRF \etal}
 & \multicolumn{4}{c}{Re-projection}
 & \multicolumn{3}{c}{Latent}
 & \multicolumn{2}{c}{Other}
 & Ours
 \\
 \cmidrule(lr){2-6}
 \cmidrule(lr){7-10}
 \cmidrule(lr){11-13}
 \cmidrule(lr){14-15}
 \cmidrule(lr){16-16}
 &
  \rot{NeRF~\citep{Mildenhall20eccv_nerf}} &
  \rot{FastNeRF~\citep{Garbin21iccv_FastNeRF}} &
  \rot{NeRF-W~\citep{MartinBrualla21cvpr_nerfw}} &
  \rot{BARF~\citep{Lin21iccv_BARF}} &
  \rot{DietNeRF~\citep{Jain21iccv_DietNeRF}} &
  \rot{GRF~\citep{Trevithick21iccv_GRF}} &
  \rot{IBRNet~\citep{Wang21cvpr_IBRNet}} &
  \rot{PixelNeRF~\citep{Yu21cvpr_pixelNeRF}} &
  \rot{MVSNeRF\citep{Chen21iccv_MVSNeRF}} &
  \rot{CodeNeRF~\citep{Jang21iccv_CodeNeRF}} &
  \rot{NeRF-VAE~\citep{Kosiorek21icml_NeRF_VAE}} &
  \rot{LFN~\citep{sitzmann2021lfn}} &
  \rot{NRW~\citep{nrw}} &
  \rot{GFVS~\citep{gfvs}} &
  \rot{\lit{}} \\
\cmidrule{2-16}
{1) Real-world data}      & \y & \y & \y & \y & \y & \y & \y & \y & \y & \n & \n & \n & \y & \y & \y \\
{2) Temp. consistent}     & \y & \y & \y & \y & \y & \y & \y & \y & \y & \y & \y & \y & \n & \n & \y \\
{3) Real-time}            & \n & \y & \n & \n & \n & \n & \n & \n & \n & \n & \n & \y & \n & \n & \y \\
{4) Appearance enc.}      & \n & \n & \y & \n & \n & \n & \n & \n & \n & \n & \n & \n & \n & \n & \y \\
{5) Pose-free}            & \n & \n & \n & \y & \n & \n & \n & \n & \n & \y & \n & \n & \n & \n & \y \\
{6) Few images}           & \n & \n & \n & \n & \y & \n & \n & \y & \y & \y & \y & \y & \y & \y & \y \\
{7) Generalization}       & \n & \n & \n & \n & \n & \y & \y & \y & \y & \y & \y & \y & \n & \y & \y \\
{8) Instant new scene}    & \n & \n & \n & \n & \n & \y & \y & \y & \y & \n & \y & \n & \n & \y & \y \\
{9) Global latent}        & \n & \n & \n & \n & \n & \n & \n & \n & \n & \y & \y & \y & \n & \n & \y \\
\bottomrule
\end{tabular}
}
\vspace{-.5em}
\caption{
    \textbf{Comparison of features with prior methods.}
    The rows indicate whether each method:
    1) has been demonstrated on real world data,
    2) is designed to provide coherence between nearby views,
    3) performs inference in real-time,
    4) handles images with varying appearance in the same scene,
    5) works without known camera poses at test time,
    6) works with very sparse input images,
    7) generalizes from a training set to novel scenes,
    8) captures novel scenes in a feed-forward step, and
    9) learns a global latent scene representation.
\vspace{-5mm}
}
\label{tab:baselines}
\end{table}

%% file: figs/architecture.tex
\begin{figure*}[t]
\centering
\includegraphics[clip, trim=15mm 10mm 23mm 17mm, width=0.92\textwidth]{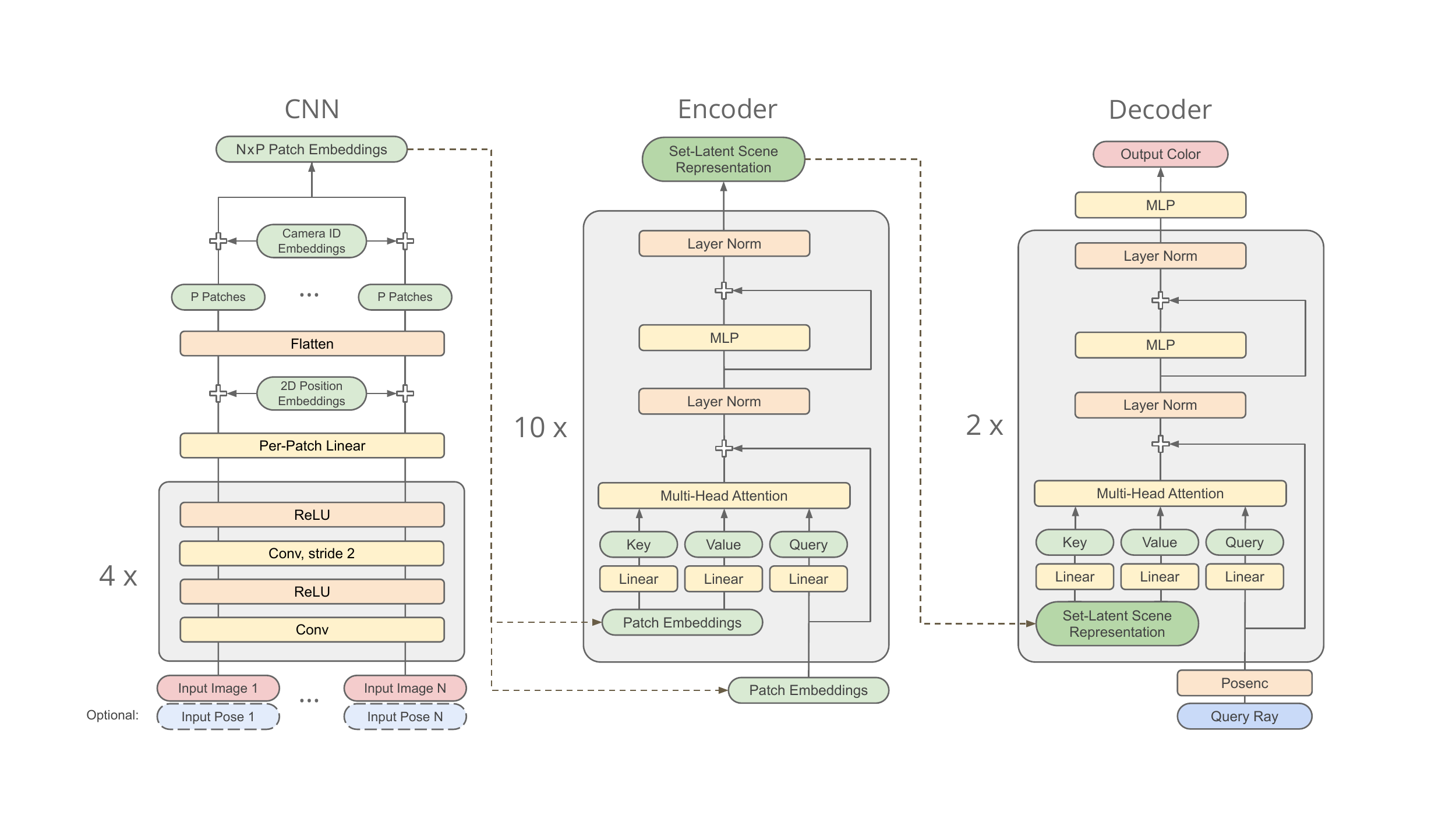}
\vspace{-4mm}
\caption{
\textbf{Network architecture} --
    Given a set of posed (\lit{}) or unposed (\uplit{}) RGB images, a CNN extracts patch features, which are processed by an encoder transformer, leading to the \emph{set-latent \gsr{}}.
    Novel views are then rendered by attending into the \gsr{} for a given ray pose, from which an image can then be rendered.
    Details in \cref{sec:method,sec:app:architecture}.
\vspace{-3mm}
}
\label{fig:architecture}
\end{figure*}

%% file: tex/3-method.tex
\section{Method}
\label{sec:method}
 
Our model receives as input an unordered \textit{collection} of optionally posed images $\{ \image_\iImage~{\in}~\real^{\H \times \W \times 3}, \camera_\iImage~{\in}~\text{SE}(3), K_\iImage \in \real^{3\times3} \}$ of the \textit{same} scene, with extrinsic camera matrix $\camera_\iImage = [R_\iImage|t_\iImage]$ and intrinsic camera matrix $K_\iImage$.
The model encodes this collection into a \textit{set} representation of cardinality $\nCodes$ via an encoder $\encoder$ with trainable parameters~$\pars_\encoder$:
\begin{equation}
\{\code_\iCode \in \real^\dCode \} = \encoder_{\pars_\encoder}( \:
    \backbone_{\pars_\backbone}(\{\image_\iImage, \camera_\iImage, K_\iImage\}))
\label{eq:encoder}
\end{equation}
In more details, our encoder first processes the images via a shared \backbone{} backbone (\cref{fig:architecture}, left), and then aggregates features into a set of flat patch embeddings.
This set of embeddings is then processed by the encoder transformer that generates the codes~$\scene$ as output.
Note that $\scene$ fully encodes a specific 3D scene as observed by the corresponding set of images, hence it is our \emph{set-latent scene representation}.

This representation is queried by the decoder transformer $\decoder$ with trainable parameters~$\pars_\decoder$ to generate the pixel color for a view ray $\ray = (\rayOrigin, \rayDir)$ with origin $\rayOrigin$ and direction $\rayDir$:
\begin{equation}
    \C(\ray) = \decoder_{\pars_\decoder}(\ray \given \{\code_\iCode\})
\end{equation}
\label{sec:method:input}
By convention, we assume $\image_0$ to be the \textit{canonical} image, hence queries $\ray$ are expressed in the coordinates of~$\camera_0$, and, whenever camera information is not available, we assume the canonical camera pose to be identity.
If pose information is known, images within a scene are randomly shuffled, hence the reference frame is randomly chosen both at train and test time.
Given a collection of images $\{\image_{\iScene,\iImage}\}$ from different scenes indexed by $\iScene$, we train all parameters end-to-end via a reconstruction loss on novel-view synthesis: 
\begin{equation}
    \argmin_{\pars=[\pars_\backbone,\pars_\encoder, \pars_\decoder]} \:\:
    \sum_\iScene
    \expect_{\ray \sim \image_{\iScene,\iImage}^\text{gt}}
    \|
    \C(\ray)
    - \image_{\iScene,\iImage}^\text{gt}(\ray)
    \|_2^2
\label{eq:loss}
\end{equation}
As both inputs and outputs of the model during training are 2D images, there is no explicit 3D supervision.
We now describe the internals of our architecture in details, as illustrated in \cref{fig:architecture}.

\newcommand{\FeatureEncoding}{\Feature^\posenc}
\newcommand{\FeatureCNN}{\Feature^\backbone}

\paragraph{Pose information}
We start by concatenating along the channel dimension the positional encoding $\posenc$~\citep{transformer,Mildenhall20eccv_nerf} of camera origin with $\numoctaves_\rayOrigin$ octaves and camera direction with $\numoctaves_\rayDir$ octaves, as seen from the currently selected canonical camera $\camera_0$ to produce:
\begin{align}
\label{eq:encoding}
\posenc(\ray) &= \posenc(\camera^{-1}_0 \cdot \rayOrigin,\numoctaves_\rayOrigin)
\concat \posenc(\camera^{-1}_0 \cdot \rayDir,\numoctaves_\rayDir)
\\
\FeatureEncoding_\iImage(\ray) &= \image_\iImage(\ray) \concat \posenc(\ray)
\end{align}
Overall, the output of this stage will have dimensionality~$|\FeatureEncoding_\iImage|=(\H,\,\W,\,\C)$ where $\C{=}(3+6\numoctaves_\rayo+6\numoctaves_\rayd)$ or $\C{=}3$, depending on whether camera pose information is provided or not.
Following \citep{Mildenhall20eccv_nerf}, we perform sine and cosine encoding independently for each of the 3D dimensions (\ie, $6\numoctaves$).

\paragraph{CNN -- \cref{fig:architecture} (left)}
A shared convolutional neural network extracts patch features while scaling the spatial dimensions of the input images by a factor of $\patchsize{=}16$, hence mapping the shape from~$(\H,\,\W,\,\C)$ to~$(\h=\nicefrac{\H}{\patchsize},\,\w=\nicefrac{\W}{\patchsize},\,\c)$:
\begin{align}
\FeatureCNN_\iImage &= \text{CNN}_{\pars_\backbone}(\FeatureEncoding_\iImage) + \bias(\pars)
\end{align}
\label{sec:method:cnn}
We add a single globally learned 2D position embedding $\bias(\pars)~{\in}~\real^{\h \times \w \times \c}$ to all patches of all images, to allow the model to retain the position of each patch in the corresponding image.
The spatial dimensions are then flattened to obtain $\Featurehat^{\backbone}_\iImage \in \real^{\h \w, \c}$. 
Further, one of two learned camera id embeddings $\bias_{\iImage=0}(\pars),\,\bias_{\iImage\neq0}(\pars)~{\in}~\real^{\c}$ are added to all patches of the canonical camera, and all remaining cameras, respectively.
This is to allow the model to distinguish the reference camera from all others:
\begin{equation}
\Feature_{\iImage, \patchind}
= \Featurehat^{\backbone}_{\iImage,\patchind} + 
\begin{cases}
    \bias_{\iImage=0}(\pars), \;\text{if } i=0,\\
    \bias_{\iImage\neq0}(\pars), \;\text{if } i\neq0,
\end{cases}
\!\!\!\!\!
\forall \patchind \in \{1,..,\h\w\}
\label{eq:bias}
\end{equation}
All features together form (unsorted) sets of \textit{patch embeddings} of cardinality~$|\{\patchFeature_\iPatches\}| = \nImages \h \w$, each of size $\real^{\C}$:
\begin{equation}
    \{\patchFeature_\iPatches\}
    =
    \bigcup_{\iImage=1}^\nImages
    \bigcup_{\patchind=1}^{\h\w}
    \: \Feature_{\iImage, \patchind}
\end{equation}
\input{tables/quantitative}
% ------------------------------------------------------------------------------

\paragraph{Encoder Transformer -- \cref{fig:architecture}~(center)}
We pass the patch embeddings through a standard transformer~\citep{transformer} that alternates between attention to all tokens, and small MLP blocks:
\begin{equation}
\scene = \encoder_{\pars_\encoder}(\{\patchFeature_\iPatches\})
\end{equation}
This network integrates information of the scene at a global scale by attending across patches and input images to infer a 3D scene representation.
Note that crucially, the set-latent scene representation scales in size with the amount of input information rather than being fixed-size~\cite{Jang21iccv_CodeNeRF, Kosiorek21icml_NeRF_VAE, sitzmann2021lfn}.

% ------------------------------------------------------------------------------

\paragraph{Decoder Transformer -- \cref{fig:architecture}~(right)}
The decoder $\decoder$ is also a transformer.
The main difference with respect to the encoder is that the ray corresponding to the pixel to be rendered is used as the \emph{query} for the multi-head attention mechanism, while key and value are computed from the scene representation $\scene$ in all layers.
In other words, the decoder learns to \textit{attend} to the most relevant subset of features in the scene representation to calculate the output color.
Note that the query ray position is encoded as in \cref{eq:encoding}.
The output of the decoder is finally passed to a small 2-layer MLP to compute the pixel color.

% \vspace{-1mm}
\subsection{Training and Inference}
\vspace{-1mm}

During training, all components are simply trained end-to-end using an L2 reconstruction loss on novel views, see \cref{eq:loss}.
We provide further training details in the \cref{sec:app:trainingdetails}.
For inference, input images are encoded into a scene representation once, which can then be used to render an arbitrary number of novel views (\eg, for a video).
Note that this methodology is designed to maximize inference efficiency, as the \emph{3D reasoning} is executed only \emph{once} in the encoder transformer, independent of the number of novel views to be rendered.
For downstream tasks, the encoder for the scene representation can be pre-trained and either frozen, or fine-tuned along with a new decoder for a novel task (e.g. semantic segmentation - \cref{sec:exp:semantic}).

%% file: tables/quantitative.tex
\begin{table}[t]
\centering
\small
\setlength{\tabcolsep}{3pt}
\def\arraystretch{1}
\resizebox{\linewidth}{!}{
\scriptsize
\begin{tabular}{lcccccc}
	\toprule
	& \multicolumn{3}{c}{NMR} & \multicolumn{3}{c}{Multi-ShapeNet} \\
	\cmidrule(lr){2-4} \cmidrule(lr){5-7}
	& $\uparrow$\,PSNR
	& $\uparrow$\,SSIM
	& $\downarrow$\,LPIPS
	& $\uparrow$\,PSNR
	& $\uparrow$\,SSIM
	& $\downarrow$\,LPIPS \\
    \cmidrule(lr){2-4} \cmidrule(lr){5-7}
	{LFN~\cite{sitzmann2021lfn}}
	& 24.95 & 0.870 & - & 14.77 & 0.328 & 0.582 \\
	{PN~\cite{Yu21cvpr_pixelNeRF}}
	& \sbest 26.80 & \sbest 0.910 & \sbest 0.108 & \sbest 21.97 & \sbest 0.689 & \best\bf{0.332} \\
	{\lit{} (Ours)}
	& \best\bf{27.87} & \best\bf{0.912} & \best\bf{0.066}  & \best\bf{23.41} & \best\bf{0.697} & \sbest 0.369 \\
	\bottomrule
\end{tabular}
}
\caption{
\textbf{Quantitative results} --
Evaluation of new scene novel-view synthesis on simple and challenging datasets.
NMR baselines provided by the authors (LPIPS for LFN not available).
\lit{} is best across datasets and metrics, with the exception of LPIPS on \msn{}.
}
\vspace{-4mm}
\label{tab:quantitative}
\end{table}

%% file: tex/4-exps.tex
\input{tables/benchmark}

\vspace{-1mm}
\section{Experimental Results}
\label{sec:exp}
\vspace{-1mm}

We run a series of experiments to evaluate how well \lit{} performs novel view synthesis in comparison to previous works and to study how each component of the network architecture contributes to the results.
In all experiments, all methods are pre-trained on a dataset of images from one set of scenes and then tested on images from novel scenes.
Unless stated otherwise, we train \lit{} with the same model architecture, hyper parameters, and training protocols for all experiments across all datasets.
Synthesized images are evaluated with PSNR, SSIM~\cite{ssim}, and LPIPS~\cite{lpips}~(VGG).

\subsection{Datasets}
\label{sec:datasets}

\paragraph{Neural 3D Mesh Renderer Dataset (NMR)~\citep{kato2018neural}}
NMR has been used in several previous studies.
It consists of ShapeNet~\cite{chang2015shapenet} objects rendered at 24 fixed views.
The dataset is very simple, and likely does not provide good evidence for the applicability in real-world settings.

\paragraph{MultiShapeNet (\msn{})}
We therefore propose this significantly more challenging dataset rendered using photo-realistic ray tracing~\cite{greff2021kubric}.
Each scene has 16-31 ShapeNet objects dropped at random locations within an invisible bounding box.
We further sample from 382 complex HDR backgrounds and environment maps.
Viewpoints are selected by uniformly sampling within a half-sphere shell.

This dataset is significantly more difficult than NMR for several reasons:
1) images are rendered photorealistically,
2) scenes contain many different types of objects in complex, tightly-packed configurations,
3) the nontrivial background maps hamper the model's abilities to segment the objects (just like in the real world), and
4) the randomly sampled viewpoints prevent models from overfitting, unlike NMR where the same cameras and canonically oriented objects are used.
The dataset is available at \litwebsite{}.

\paragraph{Street View}
\label{sec:sv}
This is a new dataset created from real-world Street View data~\cite{GoogleStreetView}.
The training dataset contains 5.5\,M scenes from San Francisco, each with 10 viewpoints.
This dataset is very challenging,
since the input viewpoints are typically several meters from the reference viewpoints,
the cameras are often arranged in straight lines (along the car trajectory, meaning there is not a lot of diversity in viewing angles),
the scenes contain challenging geometry (\eg, trees and thin poles) and dynamic elements (\eg, moving vehicles and pedestrians),
the cameras have fisheye distortion and rolling shutter (making reprojection-based methods such as~\citep{Yu21cvpr_pixelNeRF, Trevithick21iccv_GRF, Wang21cvpr_IBRNet} inapplicable), and finally,
different images often contain different exposure and white-balance settings.

%%%%%%%%%%%%%%%%%%%%%%%%%%%%%%%%%%%%%
\subsection{Comparisons to Baselines}
\label{sec:baselines}

%%%%%%%%%%%%%%

In the first set of experiments, we compare the results of \lit{} to previous methods for novel view synthesis.
For these experiments, we restrict our tests to the NMR and \msn{} datasets, since they provide an opportunity for \emph{direct comparisons} to recently published methods in a synthetic setting with known ground truth.
We run \lit{} with the same parameters on both datasets. 

%%%%%%%%%
\paragraph{Baselines}
We provide comparisons to two baselines that share aspects of our approach.
The first baseline is PixelNeRF~\citep{Yu21cvpr_pixelNeRF}, which also pretrains a network to produce features from images.
The second baseline is Light Field Networks (LFN)~\citep{sitzmann2021lfn}, which like our method uses a light field formulation rather than volumetric rendering.
These methods have demonstrated favorable quantitative and qualitative results in comparison to other previous work~\citep{Trevithick21iccv_GRF, Sitzmann19neurips_srn, Niemeyer20cvpr_DVR, softras}.

%%%%%%%%
\paragraph{Quantitative Results -- \cref{tab:quantitative,tab:benchmark}}
\lit{} provides the best image quality on NMR, outperforming LFN and PixelNeRF.
In terms of computational performance (\cref{tab:benchmark}), both PixelNeRF and \lit{} are fast to encode new scenes, while LFN requires slow optimization of the scene embedding.
Once a scene is encoded, both LFN  and \lit{} render new frames at interactive rates, while PixelNeRF is significantly slower due to the volumetric rendering.
\lit{} is the only method fast at both scene encoding and novel-view generation, making it suitable for the practical application of rendering videos of novel scenes at interactive rates.
See \cref{sec:app:benchmarkdetails} further details.
We note that SRT is comparably small for a transformer: its 74\,M parameters consist of 23\,M for CNN, 47\,M for Encoder, and 4\,M for Decoder.
For comparison, ViT~\citet{vit} has 86\,-\,632\,M parameters.

On the \msn{} dataset, \lit{} outperforms LFN and PixelNeRF in terms of PSNR and SSIM.
In terms of LPIPS, PixelNeRF outperforms \lit{}, which we attribute to the fact that \lit{} tends to blur regions of uncertainty.
In a separate experiment, we tried increasing the latent dimension of LFN (from 256 to 1024) to account for more complex scenes, but this did not increase PSNR on novel scenes.

%%%%%%%%%%%%%%
\input{figs/shapenet}
\input{figs/multishapenet}
\paragraph{Qualitative Results -- \cref{fig:shapenet,fig:multishapenet}}
The results show that our method outperforms both baselines on the NMR dataset.
While all methods are able to synthesize high quality images for nearby views (\cref{fig:shapenet}, center), PixelNeRF renders degrade particularly quickly as the camera moves away from the input views (\cref{fig:shapenet}, bottom).
LFN render quality degrades similarly, and it is also blurry for more complex objects.
In contrast, \lit{} provides similar performance for a wide range of views.
The \msn{} dataset (\cref{fig:multishapenet}) with its complex scenes clearly demonstrates the limits of LFN's \emph{single-latent} scene representation.
While PixelNeRF results in detailed reconstructions for target views close to the inputs (\cref{fig:multishapenet} middle), \lit{} is the only method that allows for good reconstructions of far-away target views (\cref{fig:multishapenet} bottom).

%%%%%%%%
\paragraph{Model Inspection -- \cref{fig:attention}}
\label{sec:exp:attention}
Further insight can be gleaned by investigating the attention weights of the encoder and decoder transformers.
We can see from \cref{fig:attention} (top) that the patch with the small chair (marked with a green box) mostly attends into the same chair in the other input images, even when the chair is facing different directions.
We further noticed that all patches additionally attend into a subset of patches positioned along the edges of the images (bottom corners of the first input image).
We believe that the encoder learns to store global information in those specific patches.
 
A glance at the decoder supports this conjecture.
When rendering a ray of the chair from a novel direction, the first attention layer only attends into the bottom edge of all input views, while the second layer attends into similar patches as the encoder, and into the bottom corners of the first input image.
This learned 2-stage inference in the decoder appears to be crucial, as we noticed in prior experiments that a 1-layer decoder is significantly worse, while more than two layers do not lead to significant gains.
The model appears to have learned a \emph{hybrid global\,/\,local} conditioning pattern through back propagation, without explicit geometric projections.

%%%%%%%%%%%%%%%%%%%%%%%%%%%%%%%%%%%%%
\subsection{Ablation Studies}
\label{sec:ablations}

We now perform a series of ablation studies on the \msn{} dataset, where we remove or substitute the main components of \lit{} and measure the change in performance to evaluate individual contributions of different design decisions.

%%%%%%
\paragraph{No Encoder Transformer}
Retraining the system without an encoder transformer (\ie, the decoder attends directly into the output of the CNN) provides a significant drop from 23.41 to 21.64 PSNR.
This suggests that the encoder transformer adds crucial capacity for inference, and that the choice to move compute from the decoder (which should be as small and fast as possible) to the encoder is a feasible design decision.
Note that its computational overhead is negligible in most practical settings, as it is only a single feed-forward step, and only run once per novel scene, independent of the number of rendered frames.

%%%%%%
\paragraph{Flat Latent Scene Representation}
One of the main novelties of \lit{} is the \emph{set-latent} scene representation.
We investigate the more commonly used flat latent by feeding the mean patch embedding of the encoder transformer into a large 8-layer MLP and dropping the decoder transformer.
This architecture leads to a significant drop from 23.41 down to 20.88 in PSNR, showing the strength of a large set scene representation along with an attention-decoder.

%%%%%%
\paragraph{Volumetric Rendering}
As an alternative to the light field formulation, we also investigate a volumetric parametrization.
To this end, we simply query the decoder not with rays, but rather with 3D points, followed by the volumetric rendering~\citep{Mildenhall20eccv_nerf}.
For simplicity, we do not inject viewing directions, and only use a single \emph{coarse} network.
We call this variation \vlit{}.
While this variation leads to an explicit 3D volume with easy-to-visualize depth maps, and results are visually comparable to \lit{}, it is to be mentioned that the \vlit{}'s decoding is slower by a theoretical factor of 192$\times$, the number of samples per ray, making inference time of this variation more similar to existing volumetric methods~\citep{Yu21cvpr_pixelNeRF}.
It is notable that our model architecture can be trained in both setups without further changes to architecture or hyper-parameters.
See details in \cref{sec:app:volumetric}.

%%%%%%%%%%%%%%%%%%%%%%%%%%%%%%%%%%%%%
\subsection{Robustness Study}
\label{sec:exp:sensitivity}
\label{sec:exp:unposed}
\input{figs/attention}

While perfect pose information is available in synthetic setups, real-world applications often depend on estimated camera poses~(\eg, \citep{colmap}), which is slow and often contains errors~\citep{Lin21iccv_BARF}.
We therefore study \lit{} and baselines in a noisy camera pose regime.
To this end, we follow~\citet{Lin21iccv_BARF} and synthetically perturb all input camera poses $\camera_\iImage$ for $\iImage \neq 0$ (all but the reference camera, see \cref{sec:method:input}) with additive noise $\delta \camera_\iImage = \mathcal{N}(0, \sigma)$ to various degrees.
For each value of $\sigma$, we retrain all models from scratch to allow them to adjust to the noise.
To allow PixelNeRF to distinguish the reference input camera $\camera_0$ from the others, we add learned camera identity embeddings to the output of the CNN, see \cref{sec:method:cnn}.

PixelNeRF heavily relues on accurate camera poses to perform projections from 3D volumes into 2D images during rendering.
As expected, \cref{plot:camnoise} shows that the performance of this method degrades sharply even with small amounts of noise.
LFN does not scale to the \msn{} dataset (see \cref{fig:multishapenet}), and we note that minor amounts of noise actually increase PSNR due to a regularization effect.
As LFN depends on test-time optimization for the scene latent, it is not robust to noisy camera positions.

In contrast, \lit{} handles noisy poses much more gracefully.
Taking it to the extreme, {\bf our method even works with fully un-posed imagery, still outperforming both baselines}.
We call this model variation without test-time poses \uplit{}.
Further inspection shows that \uplit{} is not only trivially using the first input image (for which the pose is known by definition, see \cref{sec:method:input}) but the model is in fact learning to use all input cameras, even the un-posed ones.
This is evidenced by inspection of attention patterns, and by the fact that \uplit{} significantly outperforms \lit{} with a single input view only, see \cref{sec:app:uplit}.

%%%%%%%%%%%%%%%%%%%%%%%%%%%%%%%%%%%%%
\input{plots/camnoise}
\input{figs/sv}
\subsection{Applications}

In this section, we investigate the use of \lit{} in applications.
For these experiments, we use the Street View dataset, which contains images of real-world outdoor environments.
To compensate for variations in appearance and exposure present in this dataset, we augment the network with an appearance encoder, see \cref{sec:app:appearance} for details.

\paragraph{View Interpolation}
The goal in this application is to interpolate images captured in Street View panoramas to provide smooth video transitions between them.
This application requires not only accurate view synthesis, but also temporal coherence between nearby views.
We show a representative sample of interpolated views in \cref{fig:sv}.
Although the renders are blurrier than the input views, our results demonstrate that \lit{} scales to complex real-world scenes with nontrivial camera pose distributions.
It also shows that the model learns enough 3D scene information to render novel views far away from the inputs (last column).
We provide videos and further results showing temporal coherence in render videos in \cref{sec:app:viewinterpolation}.

\paragraph{Semantic Segmentation}
\label{sec:exp:semantic}
Our goal in this application is to predict dense semantics for novel views of outdoor scenes.
Rather than synthesizing the color images for novel views and then employing a 2D image segmentation network on them, we show that \lit{}'s scene representation trained on RGB can be leveraged more directly in new domains.
Once \lit{} has been trained for the RGB reconstruction task, we freeze the encoder, and train a new decoder transformer to synthesize semantic segmentation images from the frozen \gsr{} directly.
The semantic decoder has the same network architecture as the color decoder, except for the final output layer that is changed from 3 RGB channels to 46 semantic classes.
We train the semantic decoder using the standard multi-class cross-entropy loss in a semi-supervised setup, see \cref{sec:app:semanticsegmentation}.

Example semantic segmentation results are shown in \cref{fig:sv}.
These results suggest that the \gsr{} learned through a color reconstruction loss contains enough information about the scene to allow semantic reasoning.
Though this is just one example of many possible downstream tasks, it is a significant finding that \lit{} has learned a scene representation useful for a non-trivial application.

%% file: tables/benchmark.tex
\begin{table}[t]
    \centering
    \small
    \setlength{\tabcolsep}{3pt}
    \def\arraystretch{1}
    \resizebox{\linewidth}{!}{
    \begin{tabular}{lccc}
		\toprule
		& {LFN~\cite{sitzmann2021lfn}} 
		& {PN~\cite{Yu21cvpr_pixelNeRF}}
		& {\lit{}} \\
        \cmidrule(lr){2-4}
		{Model size (\# parameters)}   & 105\,M             & 28\,M              & 74\,M \\
		\midrule
		{a) Scene encoding time}          & $\sim$\,100\,s     & \best\bf{0.005\,s} & \sbest 0.010\,s  \\
		{b) Image rendering speed}        & \best\bf{192\,fps} & 1.3\,fps           & \sbest 121\,fps \\
		{c) New-scene video rendering}    & $\sim$\,100\,s     & \sbest 75.5\,s     & \best\bf{0.182\,s} \\
		\bottomrule
	\end{tabular}
    }
    \caption{
    \textbf{Computational performance} --
        Model size, and
        a) time required to encode a scene,
        b) frame rate for individually rendering 100 frames after encoding,
        c) application: total time to encode a novel scene and render a video of 100 frames.
        \lit{} is orders of magnitudes faster for real-world use.
        See \cref{sec:app:benchmarkdetails} for details.
    }
    \label{tab:benchmark}
\vspace{-2mm}
\end{table}

%% file: figs/shapenet.tex
\newcommand\nmrpica[1]{
\includegraphics[clip,trim=0 7px 0 0px,width=0.2\columnwidth]{imgs/nmr/#1-1}}
\newcommand\nmrpicb[1]{
\includegraphics[clip,trim=0 9px 0 12px,width=0.2\columnwidth]{imgs/nmr/#1-2}}
\newcommand\nmrpicc[1]{
\includegraphics[clip,trim=0 10px 0 12px,width=0.2\columnwidth]{imgs/nmr/#1-3}}

\def\nmrvpad{-2mm}

\begin{figure}[t]
    \centering 
    \setlength{\tabcolsep}{-2pt}
    \def\arraystretch{1}
    \begin{tabular}{ccccc}
    \multicolumn{1}{c}{Input} &
    \multicolumn{1}{c}{LFN~\cite{sitzmann2021lfn}} &
    \multicolumn{1}{c}{PN~\cite{Yu21cvpr_pixelNeRF}} &
    \multicolumn{1}{c}{\lit{} (Ours)} &
    \multicolumn{1}{c}{Target}
    \\
    \nmrpica{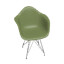} &
    \nmrpica{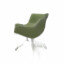} &
    \nmrpica{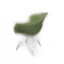} &
    \nmrpica{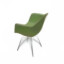} &
    \nmrpica{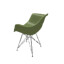}
    \\ [\nmrvpad]
    \nmrpicb{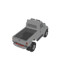} &
    \nmrpicb{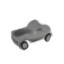} &
    \nmrpicb{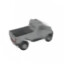} &
    \nmrpicb{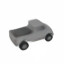} &
    \nmrpicb{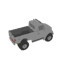}
    \\ [\nmrvpad]
    \nmrpicc{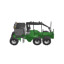} &
    \nmrpicc{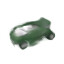} &
    \nmrpicc{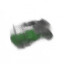} &
    \nmrpicc{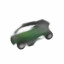} &
    \nmrpicc{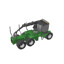}
    \\ [\nmrvpad]
    \end{tabular} 
    \caption{
        \textbf{Qualitative NMR results} --
        While PixelNeRF has high-quality for target views \emph{close} to the input (middle), results quickly degrade when further away (bottom).
        The quality of \lit{} renders is much more consistent and outperforms both baselines.
        More results are available at \litwebsite{}.
        \vspace{-4mm}
    } 
    \label{fig:shapenet}
\end{figure}

%% file: figs/multishapenet.tex
\newcommand\mspic[1]{
\includegraphics[width=0.077\linewidth]{imgs/ms/#1}
}

\def\msvpad{-1mm}

\begin{figure*}[t]
    \centering 
    \setlength{\tabcolsep}{-0.5mm}
    \def\arraystretch{2}
    \begin{tabular}{ccccc@{\hskip 0.5mm}c@{\hskip 0.1mm}c@{\hskip 0.1mm}c@{\hskip 0.5mm}c@{\hskip 0.1mm}c@{\hskip 0.5mm}c@{\hskip 0.5mm}c}
    \multicolumn{5}{c}{Input Images} &
    \multicolumn{3}{c}{Perfect Input Pose} &
    \multicolumn{2}{c}{Noisy Pose $\sigma\,$=\,0.1} &
    \multicolumn{1}{c}{No Pose} &
    \multicolumn{1}{c}{Target}
    \\[-2mm]
    \cmidrule(lr){1-5} \cmidrule(lr){6-8} \cmidrule(lr){9-10} \cmidrule(lr){11-11} \cmidrule(lr){12-12}\\[-11mm]
    \multicolumn{5}{c}{} &
    \multicolumn{1}{c}{LFN\citep{sitzmann2021lfn}} &
    \multicolumn{1}{c}{PN~\citep{Yu21cvpr_pixelNeRF}} &
    \multicolumn{1}{c}{\lit{}} &
    \multicolumn{1}{c}{PN~\citep{Yu21cvpr_pixelNeRF}} &
    \multicolumn{1}{c}{\lit{}} &
    \multicolumn{1}{c}{\uplit{}} &
    \multicolumn{1}{c}{}
    \\ [\msvpad]
    \mspic{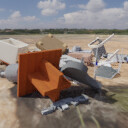} &
    \mspic{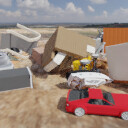} &
    \mspic{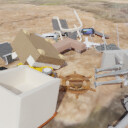} &
    \mspic{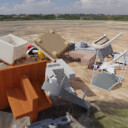} &
    \mspic{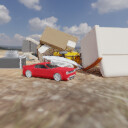} &
    \mspic{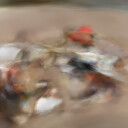} &
    \mspic{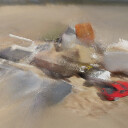} &
    \mspic{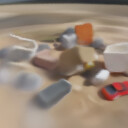} &
    \mspic{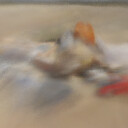} &
    \mspic{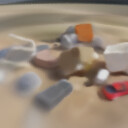} &
    \mspic{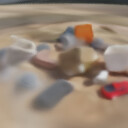} &
    \mspic{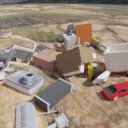}
    \\ [\msvpad]
    \mspic{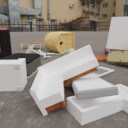} &
    \mspic{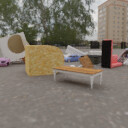} &
    \mspic{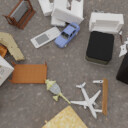} &
    \mspic{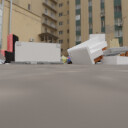} &
    \mspic{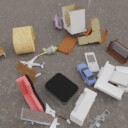} &
    \mspic{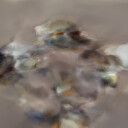} &
    \mspic{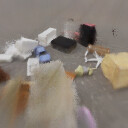} &
    \mspic{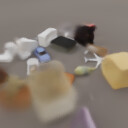} &
    \mspic{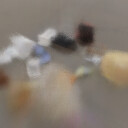} &
    \mspic{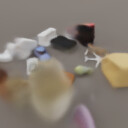} &
    \mspic{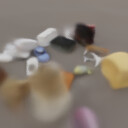} &
    \mspic{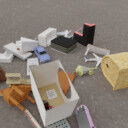}
    \\ [\msvpad]
    \mspic{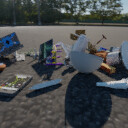} &
    \mspic{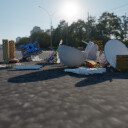} &
    \mspic{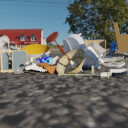} &
    \mspic{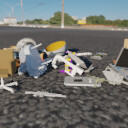} &
    \mspic{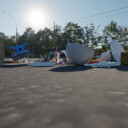} &
    \mspic{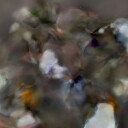} &
    \mspic{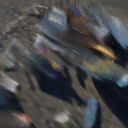} &
    \mspic{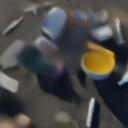} &
    \mspic{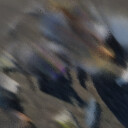} &
    \mspic{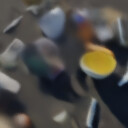} &
    \mspic{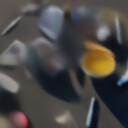} &
    \mspic{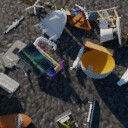}
    \\ [-3mm]
    \end{tabular} 
    \caption{
        \textbf{Qualitative results on MultiShapeNet --}
        LFN does not scale to this demanding dataset due to its global latent conditioning.
        With perfect input camera poses, PixelNeRF resolves details in the center of the scene more sharply for target views nearby the inputs (middle).
        This quickly changes for views further away (bottom), where PixelNeRF produces projection artifacts even with perfect pose, while \lit{}'s results are more coherent.
        PixelNeRF further has trouble compensating for noisy cameras, where \lit{} only experiences a mild drop in quality.
        Finally, \uplit{} is the only model that can be run without input camera poses at all (see \cref{sec:exp:sensitivity}).
    } 
    \label{fig:multishapenet}
\vspace{-3mm}
\end{figure*}

%% file: figs/attention.tex
\def\attnheight{12.5mm}

\begin{figure}[t]
    \centering
    \setlength{\tabcolsep}{1pt}
    \def\arraystretch{1}
    \small
    \begin{tabular}{r@{\hskip 1mm}cc}
    & Query & Key Patches in Input Images \\
    \cmidrule(lr){2-2} \cmidrule(lr){3-3}
    \rotatebox[origin=c]{90}{\hspace{10mm} \parbox{\attnheight}{\footnotesize \centering Encoder \\ all layers}} &
    \includegraphics[clip, trim=512px 0 0 0, height=\attnheight]{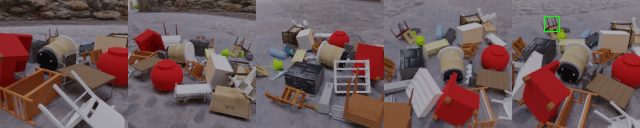} &
    \includegraphics[height=\attnheight]{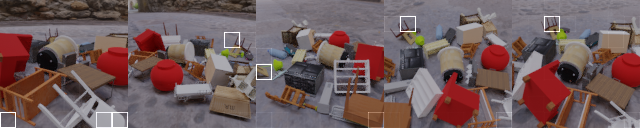}
    \\[-10mm]
    \rotatebox[origin=c]{90}{\hspace{10mm} \parbox{\attnheight}{\footnotesize \centering Decoder \\ layer 1}} &
    \includegraphics[height=\attnheight]{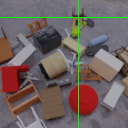} &
    \includegraphics[height=\attnheight]{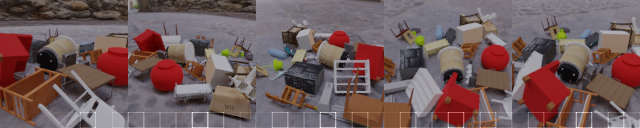} 
    \\[-10mm]
    \rotatebox[origin=c]{90}{\hspace{10mm} \parbox{\attnheight}{\footnotesize \centering Decoder \\ layer 2}} &
    \includegraphics[height=\attnheight]{imgs/attn/decoder_target} &
    \includegraphics[height=\attnheight]{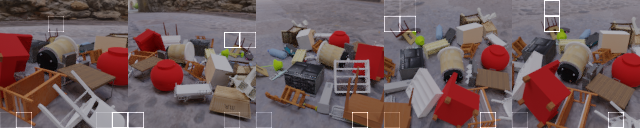}
    \\[-10mm]
    \end{tabular} 
    \caption{
    \textbf{Attention visualization} --
    Input patches that the green input patch attends into in the encoder, and the first \& second decoder layers attend into when rendering the marked query ray at the intersection of the green lines.
    The model learns to attend into the same 3D positions, and to store global information into specific tokens (along the bottom edge).
    The decoder first attends into the global patches, then into relevant 3D positions of the scene.
    \vspace{-4mm}
    }
    \label{fig:attention}
\end{figure}

%% file: plots/camnoise.tex
\begin{figure}[t]
\centering
\includegraphics[width=0.9\columnwidth]{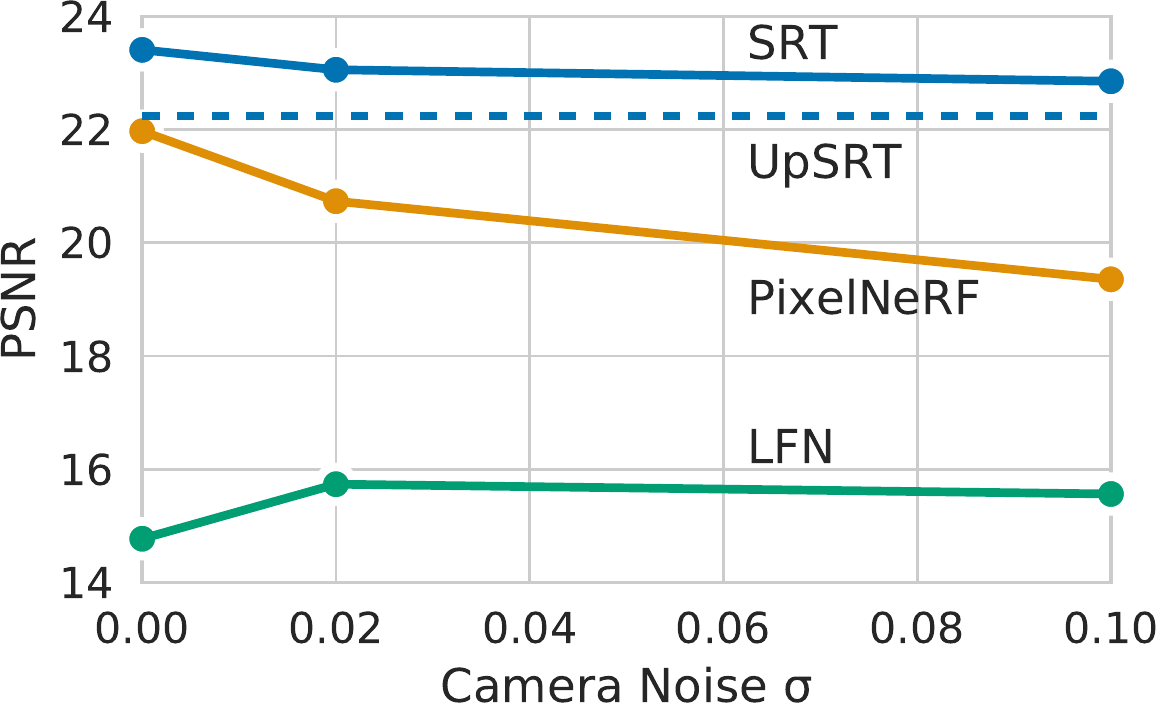}
\caption{
    \textbf{Robustness} --
    PixelNeRF~\citep{Yu21cvpr_pixelNeRF} quality drops quickly even for minor amounts of noise despite being trained for this setting.
    LFN~\citep{sitzmann2021lfn} fails to scale to \msn{} even with perfect cameras -- some noise regularizes the model and actually leads to a small increase in quality.
    \lit{} handles camera noise gracefully, outperforming competing methods even without any camera parameters (\uplit{}).
    \vspace{-5mm}
}
\label{plot:camnoise}
\end{figure}

%% file: figs/sv.tex
\newcommand\svpic[1]{
    \includegraphics[clip, trim=0 1px 0 16px,height=24mm]{imgs/sv/#1}
}

\def\svpad{0mm}

\begin{figure*}[t]
    \centering 
    \setlength{\tabcolsep}{-2pt}
    \def\arraystretch{1}
    \begin{tabular}{c@{\hskip 0.5mm}cc@{\hskip 0mm}cc}
    \multicolumn{1}{c}{Input Views} &
    \multicolumn{2}{c}{Nearby \lit{} Predictions} &
    \multicolumn{2}{c}{Far \lit{} Predictions}
    \\
    \cmidrule(lr){1-1} \cmidrule(lr){2-3} \cmidrule(lr){4-5}
    \svpic{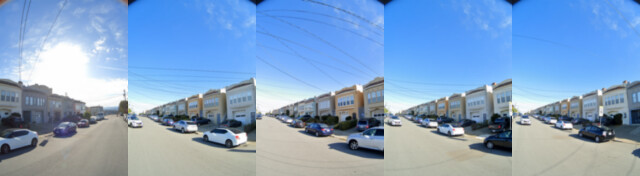} &
    \svpic{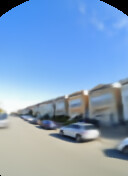} &
    \svpic{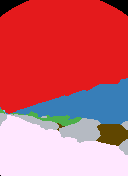} &
    \svpic{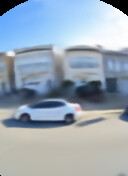} &
    \svpic{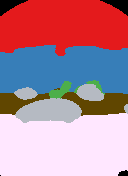}
    \\ [\svpad]
    \svpic{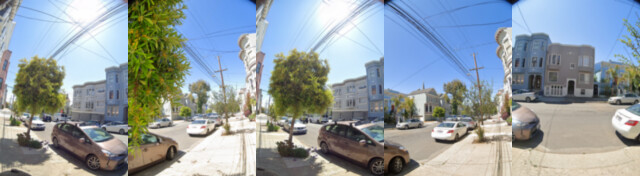} &
    \svpic{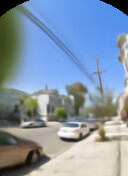} &
    \svpic{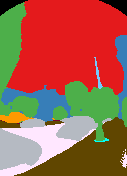} &
    \svpic{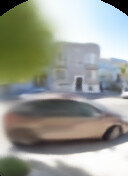} &
    \svpic{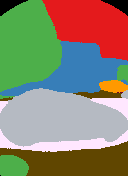}
    \\ [\svpad]
    \svpic{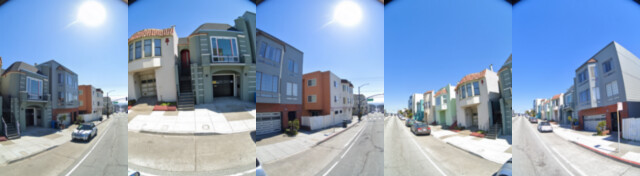} &
    \svpic{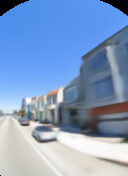} &
    \svpic{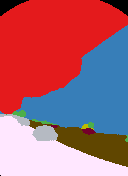} &
    \svpic{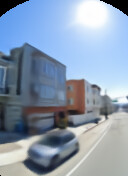} &
    \svpic{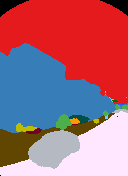}
    \\ [\svpad]
    \end{tabular} 
    \caption{
        \textbf{Qualitative results on Street View} --
        \lit{} performs reasonably on highly challenging real-world data with small and large changes in camera perspective.
        Furthermore, the scene representation contains enough information for 3D semantic scene inference.
        A comparison with NeRF-based optimization methods is provided in \cref{sec:app:nerfcomparison}.
    }
    \label{fig:sv}
\end{figure*}

%% file: tex/5-conclusion.tex
\section{Limitations}
\label{sec:limitations}

This project investigates whether transformers can learn scalable scene representations from only images without explicit geometry processing.
We here identify a few limitations, which we believe to be tackled by follow up works.

First, our results show some blurriness in images on the complex datasets, even for views close to the input cameras.
This is expected, as the model needs to learn pixel-accurate light ray transformations, which leads to uncertainty about exact positions, known to result in blurriness under an L2 loss~\citep{enet}.

Second, \lit{} is a geometry-free learning-based method, and hence will not work as well on very small datasets compared to methods with explicit geometric inductive biases.
In practice, we find that our model converges at rates similar to others, but even more training usually results in better performance.
In particular, when trained on the standard datasets reported in this paper, \lit{} outperforms LFN and PixelNeRF. 
Further investigation is required to better understand the best training protocols.

Finally, in comparisons to prior art, our model is best when input views are sparse or when camera poses are noisy or missing.
When novel views are reliably close to input views, and perfect poses are known, explicit geometric methods such as~\citet{Yu21cvpr_pixelNeRF} often provide better results, albeit with much longer inference times.
While \lit{} has been specifically designed for the sparse input scenario, future work could investigate how to better leverage nearby input views at inference time when they are available.

\section{Conclusion}
\label{sec:conclusion}

We propose \litlong{}, a model for learning scalable neural scene representations using only self-supervision from color images.
The novel encoder-decoder transformer architecture learns to render novel views without explicit geometric reasoning, and optionally without 3D image poses, yet surpassing the quality of prior art on standard benchmarks and a novel, more challenging dataset that we propose.
At the same time, \lit{} is orders of magnitudes faster than prior methods for the realistic settings such as novel-scene video generation.
This unique combination of flexibility, generality, and efficiency is well-suited for real world applications with very large datasets, and we believe that this work will inspire the community towards similar more implicit, yet scalable methods.

%% file: tex/6-acknowledgements.tex
\section{Acknowledgments}
We thank
Adam Kosiorek,
Alex Yu,
Alexander A. Kolesnikov,
Aravindh Mahendran,
Luke Barrington,
Konstantinos Rematas,
Sebastian Ebert,
Srinadh Bhojanapalli
Vickie Ye,
and
Vincent Sitzmann
for their help and fruitful discussions.

%% file: tex/8-appendix.tex
\appendix
\section{Appendix}
\label{sec:appendix}
\label{sec:app:results}

We provide a discussion of potential negative societal impact and include further details and results.
Video results are available on the project website at \litwebsite{}.

%%%%%%%%%%%%%%%%%%%%%%%%%%%%%%%%%%%%%%%%%%%%%%%%
\input{tex/7-statements}

%%%%%%%%%%%%%%%%%%%%%%%%%%%%%%%%%%%%%%%%%%%%%%%%
\subsection{Architecture, Training and Evaluation}
\label{sec:app:architecture}
\label{sec:app:trainingdetails}
\label{sec:app:trainingprotocols}

%%%%%%%%%%%%%%%%%%%%%%%%%%%%%%%%%%%%%%%%%%%%%%%%
\subsubsection*{Dataset Details}
\label{sec:app:datasets}

\paragraph{Neural 3D Mesh Renderer Dataset (NMR)~\citep{kato2018neural}}
This dataset consists of ShapeNet~\cite{chang2015shapenet} objects from the 13 largest classes, each with 24 fixed views rendered at 64$\times$64 pixels.
We use the SoftRas split~\cite{nmrdataset}, which contains 31\,k training, 4\,k validation and 9\,k test objects.
Following the protocol in~\citep{sitzmann2021lfn,Yu21cvpr_pixelNeRF}, one randomly sampled image is provided as input, and images are rendered for all other 23 views.

\paragraph{MultiShapeNet (\msn{})}
We use 1\,M scenes for training, and 10\,k for testing.
To ensure that all test set scenes contain novel objects, we split the full set of ShapeNet objects into a train and test split, and sample objects for each from the corresponding split.
Viewpoints are selected by independently and uniformly sampling 10 cameras within a half-sphere shell of radius 8-12 for each scene, with all cameras pointed at the origin.
All images are rendered at a resolution of 128$\times$128.
The task is to render novel views given 5 randomly selected input images of a novel scene.

\paragraph{Street View}
The resolution of this dataset is 176$\times$128 pixels.
During training, the task is to predict 5 randomly sampled images from the other 5 input views.
At test time, we render views from novel viewpoints.
To ensure that the evaluation scenes are new to the model, we physically separate them from the training scenes.

%%%%%%%%%%%%%%%%%%%%%%%%%%%%%%%%%%%%%%%%%%%%%%%%
\subsubsection*{\lit{} Model Details}
We use the same network architecture and hyper parameters throughout all experiments (with the exception of the appearance encoding module, see Sec.~\ref{sec:app:appearance}).
Our CNN consists of 4 blocks, each with 2 convolutional layers.
The first convolution in each block has stride~1, the second has stride~2.
We begin with 96 channels which we double with every strided convolution, and map the final activations with a 1$\times$1 convolution (\ie, a per-patch linear layer) to 768 channels.

Both encoder and decoder transformers are similar to ViT-Base~\cite{vit}: all attention layers have 12 heads, each with 64 channels, and the MLPs have 1 hidden layer with 1536 channels and GELU~\cite{gelu} activations.
The final MLP for the decoder also has a single hidden layer, but with 128 channels, with a final sigmoid operation for the RGB output.
For the positional encoding of ray origin and direction, we use $\numoctaves_\rayOrigin=\numoctaves_\rayDir=15$.

We train our model with the Adam optimizer~\cite{adam} with initial learning rate \num{1e-4} that is smoothly decayed down to \num{1.6e-5} over 4\,M training steps and a learning rate warmup phase in the first 2.5\,k steps.
We train with a batch size of 256 and for every data point, we randomly sample 8192 points uniformly across all target views for rendering and the loss computation.
We noticed that the model was still not fully converged at 4\,M training steps, and validation metrics often keep improving up to and beyond 10\,M, the maximum to which we trained the model (especially on more challenging datasets), though qualitative improvements are increasingly subtle as training progresses.

%%%%%%%%%%%%%%%%%%%%%%%%%%%%%%%%%%%%%%%%%%%%%%%%
\subsubsection*{Baseline Details}

\paragraph{LFN}
We received the LFN~\cite{sitzmann2021lfn} code through private communications with the authors.
The code was adapted to read the \msn{} dataset. 
We trained LFN for 500\,k steps on the NMR dataset with batch size 128 and using the Adam optimizer with learning rate \num{1e-4}.
After training, the weights were frozen, and only the latent codes were optimized for 2\,k epochs with batch size 64.
For \msn{}, we trained LFN using a batch size of 16 for 500\,k steps as training converged much faster and to lower loss values than with the default batch size of 128.
For the camera pose noise ablations, we trained for 350\,k steps (those models converged earlier) with batch size 16.
After training, we performed latent code optimization for 200\,k steps using batch size 16.

\paragraph{PixelNeRF}
We used the official implementation from the authors~\cite{pncode}.
PixelNerf~\cite{Yu21cvpr_pixelNeRF} was trained for 1\,M steps with the configuration from the codebase used for all experiments in the original paper.
During training, each batch contains 4 scenes of 10 images each for the \msn{} dataset, and 128 rays are sampled per scene for the loss.
Training uses Adam with learning rate \num{1e-4}.
In consultation with the authors, we applied a minor fix in the projection code, tweaking a division operation to avoid NaNs in some experiments.

%%%%%%%%%%%%%%%%%%%%%%%%%%%%%%%%%%%%%%%%%%%%%%%%
\subsubsection*{Appearance Encoding}
\label{sec:app:appearance}
\input{figs/appendix_epi}

As the Street View data contains variations in exposure and white balance, we extend our model with an appearance encoder for these experiments.
During training, we extract an appearance embedding from the respective target views and concatenate it to the input of the final 2-layer MLP after the decoder transformer, see \cref{fig:architecture}.
The appearance encoder consists of 4 blocks of 2$\times$2 mean pooling operations followed by 1x1 convolutions with 32 channels, and a ReLU activation.
Finally, we take the mean over the spatial dimensions and map the vector to 4 channels before passing it to the decoder MLP.
At test time, the appearance embedding of one of the input images can be used.
\cref{fig:app:appearance} shows the effect of the appearance embedding on the output of the model.

\input{figs/appendix_appearance}

%%%%%%%%%%%%%%%%%%%%%%%%%%%%%%%%%%%%%%%%%%%%%%%%
\subsubsection*{Semantic Segmentation}
\label{sec:app:semanticsegmentation}

We train the semantics decoder $\decoder_\text{sem}$ by minimizing the cross-entropy between softmax class predictions and ground truth class distributions $p$ for each ray / pixel:
\begin{equation}
    \argmin_{\pars_{\decoder_\text{sem}}} \:\:
    -\sum_\iScene
    \expect_{\ray \sim \image_{\iScene,\iImage}^\text{gt}}
    p(\ray) \cdot \log(\decoder_\text{sem}(\ray))
\end{equation}
The ground truth classes were generated by running a pre-trained 2D semantic segmentation model on the training views, and extracting one-hot predictions.

%%%%%%%%%%%%%%%%%%%%%%%%%%%%%%%%%%%%%%%%%%%%%%%%
\subsubsection*{Benchmark Details}
\label{sec:app:benchmarkdetails}
We measure wall-clock performance of all models on a single NVIDIA~V100.
In \cref{tab:benchmark}, we measure three different settings.
For A), we measure the onboarding time of a novel scene, \ie for PixelNeRF and \lit{}, this is the execution time of the encoder, while for LFN, it is the time to optimize the scene-latent.
Hence, LFN takes minutes while PixelNeRF and \lit{} encode the scene in 5-10\,ms.
For B), once the scene has already been onboarded, we measure the rendering FPS in a streaming / interactive setup, \ie frame by frame, without batching.
Finally, C) measures a full application scenario that combines scene encoding and video rendering.
Here, we measure the full time taken to onboard a new scene and render a 100-frame video.
Note that we allow batching in this scenario, as we assume the camera render path is known, hence the measured times are significantly lower than the sum of scene encoding and single-image rendering times.

%%%%%%%%%%%%%%%%%%%%%%%%%%%%%%%%%%%%%%%%%%%%%%%%
\subsection{Experiments and Results}

\subsubsection*{\uplit{} Model Inspection}
\label{sec:app:uplit}
\input{figs/appendix_attention}

\uplit{} works without input poses at test time -- however, the pose of the first camera is always known by definition, as all target poses are given relative to its reference frame (see \cref{sec:method:input}).
Note that this does \emph{not} mean \uplit{} needs poses -- target views must always be defined somehow, and in this case, they are simply defined relative to an arbitrary input view.
Hence, the fact that \uplit{} works and renders meaningful images by itself does not prove yet that the model is capable of using unposed imagery; the model might only be using the first input camera and ignoring all other input views.
In the following, we show that this is not the case, and that \uplit{} indeed uses all input cameras.

First, we investigate the attention patterns for \uplit{} in \cref{fig:app:attention}.
As the results show, \uplit{} is using all input images for reconstructing novel views of the scene, showing that it is capable of finding similar structures purely based on imagery, and make use of that information for more detailed scene reconstructions.
Note that in the encoder, patches in other input images attend strongly specifically into the reference camera, possibly to \emph{spread} pose information to all other input views.
That could also explain why the decoder does not attend strongly into the reference camera.

Second, we compare \uplit{} trained and tested on 5 input images with \lit{} trained and tested only on one input image.
If \uplit{} was only trivially using the first (practically posed) input image, the quality of these two models should be similar.
However, we see that \lit{} only achieves a PSNR / SSIM / LPIPS of 19.2 / 0.53 / 0.52 on \msn{} in this setup, while \uplit{} reaches much better values of 22.2 / 0.65 / 0.41, respectively.

Finally, we take a closer look at the first row in \cref{fig:multishapenet} and notice that the red car is \emph{not} visible in the first input image (leftmost column), but in other, unposed input views.
Despite this, \uplit{} is correctly reconstructing the red car at the correct 3D position, showing that the model has detected the position of objects in unposed imagery and has correctly integrated them into the \gsr{}.
\cref{fig:app:multishapenet} shows further qualitative results for \uplit{}.

%%%%%%%%%%%%%%%%%%%%%%%%%%%%%%%%%%%%%%%%%%%%%%%%
\subsubsection*{\vlit{} Volumetric Rendering}
\label{sec:app:volumetric}

We show results for \vlit{} in \cref{fig:app:sv}.
As can be seen, the render quality is comparable to that of \lit{} (see \cref{fig:sv}), with the addition that depth maps can now be acquired as well.
Similarly, \cref{fig:app:multishapenet} shows results on the \msn{} dataset.

%%%%%%%%%%%%%%%%%%%%%%%%%%%%%%%%%%%%%%%%%%%%%%%%
\subsubsection*{Robustness Study}
\label{sec:app:sensitivity}

\cref{fig:app:camnoise} visualizes the effect of camera noise in our robustness studies, see \cref{sec:exp:sensitivity}.
As can be seen, the amount of noise is subtle, yet enough to severely affect reconstruction quality of geometry-based methods as seen in \cref{plot:camnoise}.

%%%%%%%%%%%%%%%%%%%%%%%%%%%%%%%%%%%%%%%%%%%%%%%%
\subsubsection*{Temporal Consistency}
\label{sec:app:viewinterpolation}

\cref{fig:app:epi} shows EPI for a dolly motion on an \msn{} scene.
The results show that \lit{} learns a coherent 3D scene.
Note that this consistency also implies that depth maps could be extracted from our model \cite{lfdepth,yan2019depth,wang2015occlusion}, see also \cite{sitzmann2021lfn}.

%%%%%%%%%%%%%%%%%%%%%%%%%%%%%%%%%%%%%%%%%%%%%%%%
\subsubsection*{Comparison to NeRF-based Methods}
\label{sec:app:nerfcomparison}

In contrast to \lit{} and the baselines we compare to in the paper, NeRF needs costly per-scene optimization, and is known to  fail in the few-image setting. In \cref{fig:app:sv:nerf}, we compare SRT with Mip-NeRF~\cite{Barron21iccv_Mip_NeRF} (improved version of NeRF) and DietNeRF~\cite{Jain21iccv_DietNeRF} (specialized to handle few images), and see that they both fail for this highly challenging dataset with only 5 images.

\input{figs/appendix_sv}
\input{figs/appendix_nerf}
\input{figs/appendix_camnoise}
\input{figs/appendix_multishapenet}

%% file: tex/7-statements.tex
\subsection{Potential Negative Societal Impact}
\label{sec:app:statements}

\paragraph{Fakes}
This paper proposes a method to synthesize novel views of scenes observed from a small set of input views.  As in any image synthesis system, it could be used to produce images that have misleading or fake content, particularly if an adversary selected the input images.  We plan to mitigate this issue partially by deploying the algorithms only in secure servers when running them on real world datasets, and we will clearly mark all synthesized imagery as such to reduce the chance of misinterpretation.

\paragraph{Privacy}
This paper includes experiments run with a Street View dataset containing images acquired in public areas of real world outdoor scenes.  As in any image dataset, these images could have information that negatively affects personal privacy.  To mitigate these potential impacts, we follow strict privacy policies required by the owner of the data, Google, which include rules for using the data in accordance with regional laws, anonymizing the data to avoid identification of people or license plates, and retaining the data in secure storage (https://policies.google.com/privacy).

\paragraph{Fairness}
Our Street View dataset contains images captured by a car-mounted camera platform driving on public roads in San Francisco.   Long sequences of images are captured systematically with fixed view directions and regular time intervals as the car drives.  There are biases in the dataset based on which roads are driven in which city, which align with demographic or economic factors.   However, individual images are not selected by a human photographer, which might mitigate some types of bias common in other computer vision datasets.

\paragraph{Energy consumption}
The paper proposes a method to train a transformer network architecture to produce a scene representation and arbitrary novel images from a set of input images.   The training procedure requires a lot of compute cycles, and thus has a negative impact on the environment due to its energy consumption.  However, in comparison to most competing methods based on volumetric rendering methods, the inference cost is smaller (by 6-7 orders of magnitude per scene and 2-3 orders of magnitude per image).   Thus, our proposed system is the most efficient of its kind when deployed at scale (our target use case) – where the system is trained on millions of images and then run without further fine-tuning to synthesize up to trillions of novel images.

%% file: figs/appendix_epi.tex
\def\episize{0.32\linewidth}

\begin{figure}[t]
    \centering
    \setlength{\tabcolsep}{1pt}
    \def\arraystretch{1}
    \small
    \begin{tabular}{ccc}
        Target frame & \lit{} & \vlit{} \\
        \includegraphics[clip, trim=0 0 0 0, width=\episize]{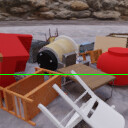} &
        \includegraphics[clip, trim=0 0 0 0, width=\episize]{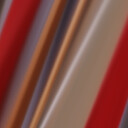} &
        \includegraphics[clip, trim=0 0 0 0, width=\episize]{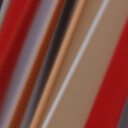}
    \end{tabular}
    \caption{
    \textbf{Epipolar Images (EPI)} --
    For a dolly camera motion, the same horizontal line of pixels marked in green (left) is stacked vertically from all video frames, producing an EPI.
    The slope of each diagonal line in an EPI corresponds to the depth of that pixel in the scene.
    \vlit{} uses a volumetric parametrization of the scene and no viewing directions.
    We can see that its EPI looks quite similar to \lit{}'s, indicating that the latter's scene representation is geometrically and temporally stable, despite a lack of theoretical consistency guarantees for our light field parametrization.
    }
    \label{fig:app:epi}
\end{figure}

%% file: figs/appendix_appearance.tex
\newcommand\aapppic[1]{
    \includegraphics[clip, trim=0 1px 0 32px,width=0.18\linewidth]{imgs/appearance/#1}
}

\begin{figure}[t]
    \centering 
    \setlength{\tabcolsep}{0mm}
    \def\arraystretch{1}
    \begin{tabular}{r@{\hskip 1mm}ccccc}
        \rotatebox[origin=c]{90}{\hspace{12mm} \small Reference} & \aapppic{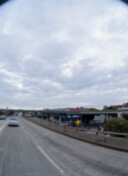} & \aapppic{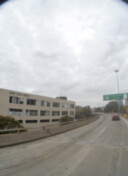} & \aapppic{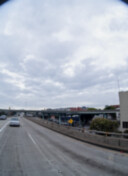} & \aapppic{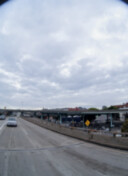} & \aapppic{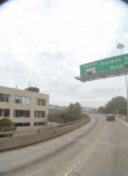} \\[-10mm]
        \rotatebox[origin=c]{90}{\hspace{12mm} \small Render} & \aapppic{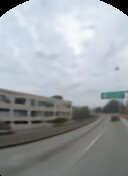} & \aapppic{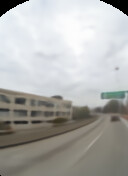} & \aapppic{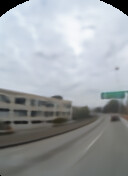} & \aapppic{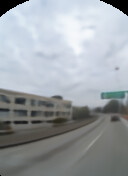} & \aapppic{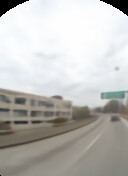} \\[-10mm]
    \end{tabular} 
    \caption{
    \textbf{Appearance embeddings} --
    In some scenes, white balance and exposure are not consistent among the input views, see the shift in colors between the reference views in the top row.
    We add a simple appearance encoder to \lit{} to handle these cases.
    Note that the appearance embedding only changes the white balance of the rendered images without changing the structure (bottom row).
    Further, it generalizes across entirely different views of the scene (\eg, leftmost column).
    The effect is similar to the appearance handling in \cite{MartinBrualla21cvpr_nerfw}, however, we do not need an expensive optimization procedure to get the appearance embedding of an image, it is a very fast feed-forward pass, and we do not need hundreds to thousands of images of the same scene to capture appearance variations.
    }
    \label{fig:app:appearance}
\end{figure}

%% file: figs/appendix_attention.tex
\def\attnheight{12.5mm}

\begin{figure}[t]
    \centering
    \setlength{\tabcolsep}{1pt}
    \def\arraystretch{1}
    \small
    \begin{tabular}{r@{\hskip 1mm}cc}
    & Query & Key Patches in Input Images \\
    \cmidrule(lr){2-2} \cmidrule(lr){3-3}
    \rotatebox[origin=c]{90}{\hspace{10mm} \parbox{\attnheight}{\footnotesize \centering Encoder \\ all layers}} &
    \includegraphics[clip, trim=512px 0 0 0, height=\attnheight]{imgs/attn/encoder_query} &
    \includegraphics[height=\attnheight]{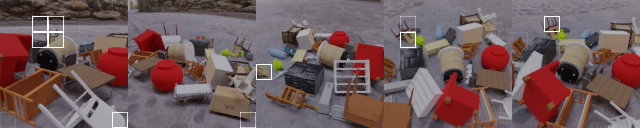}
    \\[-10mm]
    \rotatebox[origin=c]{90}{\hspace{10mm} \parbox{\attnheight}{\footnotesize \centering Decoder \\ layer 1}} &
    \includegraphics[height=\attnheight]{imgs/attn/decoder_target} &
    \includegraphics[height=\attnheight]{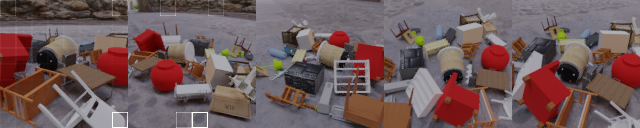} 
    \\[-10mm]
    \rotatebox[origin=c]{90}{\hspace{10mm} \parbox{\attnheight}{\footnotesize \centering Decoder \\ layer 2}} &
    \includegraphics[height=\attnheight]{imgs/attn/decoder_target} &
    \includegraphics[height=\attnheight]{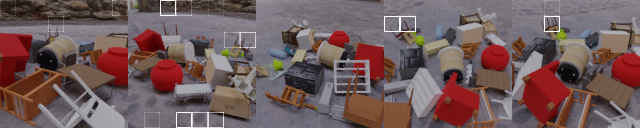}
    \\[-10mm]
    \end{tabular} 
    \caption{
    \textbf{Attention visualization for \uplit{}} --
    The attention patterns of \uplit{} resemble the ones of \lit{} (\cref{fig:attention}).
    Notably, the model attends into relevant parts of \emph{all} inputs, showing that \uplit{} has learned to use unposed imagery even in complex scenes.
    }
    \label{fig:app:attention}
\end{figure}

%% file: figs/appendix_sv.tex
\newcommand\svstpic[1]{
    \includegraphics[clip, trim=0 1px 0 16px,height=24mm]{imgs/sv_st/#1}
}

\begin{figure*}[t]
    \centering 
    \setlength{\tabcolsep}{-2pt}
    \def\arraystretch{1}
    \begin{tabular}{c@{\hskip 0.5mm}cc@{\hskip 0mm}cc}
    \multicolumn{1}{c}{Input Views} &
    \multicolumn{2}{c}{Nearby \vlit{} Predictions} &
    \multicolumn{2}{c}{Far \vlit{} Predictions}
    \\
    \cmidrule(lr){1-1} \cmidrule(lr){2-3} \cmidrule(lr){4-5}
    \svpic{1-in} &
    \svstpic{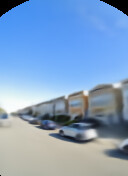} &
    \svstpic{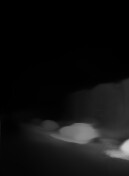} &
    \svstpic{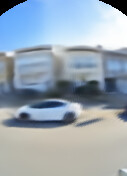} &
    \svstpic{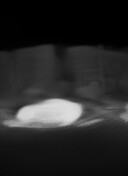}
    \\ [\svpad]
    \svpic{2-in} &
    \svstpic{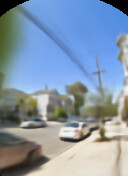} &
    \svstpic{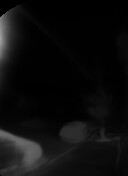} &
    \svstpic{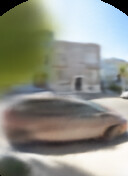} &
    \svstpic{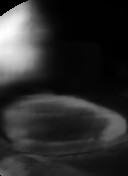}
    \\ [\svpad]
    \svpic{3-in} &
    \svstpic{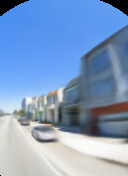} &
    \svstpic{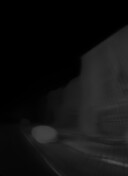} &
    \svstpic{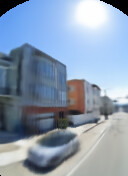} &
    \svstpic{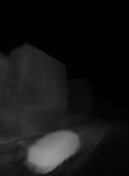}
    \\ [\svpad]
    \end{tabular} 
    \caption{
        \textbf{\vlit{} results on Street View} --
        \vlit{} has similar quality to \lit{} (see \cref{fig:sv}), but is much slower at inference due to the sampling procedure for volumetric rendering.
        However, it allows one to render depth maps, which look reasonable in most scenes.
        In the center example, the depth of the car is inaccurate as a result of the lighting variation, note that \eg the car changes color due to the metallic surface (\eg, input views 2 \vs 3), and since we did not include the viewing direction (see \cref{sec:ablations}), the model has learned to account for this in the geometry instead.
    }
    \label{fig:app:sv}
\end{figure*}

%% file: figs/appendix_nerf.tex
\newcommand\svnpic[1]{
    \includegraphics[width=0.241\linewidth]{imgs/sv_nerf/#1}
}

\def\svwidth{20mm}

\begin{figure}[t]
    \centering 
    \setlength{\tabcolsep}{1pt}
    \def\arraystretch{1}
    \begin{tabular}{cccc}
    Mip-NeRF &
    DietNeRF &
    SRT &
    Target
    \\
    12.7\,/\,0.274 &
    12.1\,/\,0.252 &
    20.5\,/\,0.699 &
    PSNR\,/\,SSIM
    \\[0.5mm]
    \svnpic{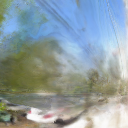} &
    \svnpic{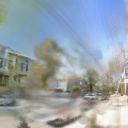} &
    \svnpic{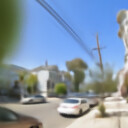} &
    \svnpic{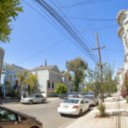}
    \\
    \svnpic{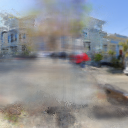} &
    \svnpic{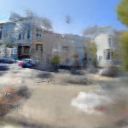} &
    \svnpic{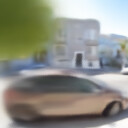} &
    \svnpic{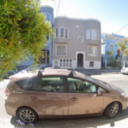}
    \end{tabular}
    \caption{Per-scene, NeRF-based optimization methods compared to \lit{} on the Street View dataset. Both Mip-NeRF~\cite{Barron21iccv_Mip_NeRF} and DietNeRF~\cite{Jain21iccv_DietNeRF} need costly per-scene optimization, and fail with only 5 input images.}
    \label{fig:app:sv:nerf}
\end{figure}

%% file: figs/appendix_camnoise.tex
\begin{figure}[t]
\centering
\includegraphics[clip,trim=0mm 35mm 35mm 96mm, width=\linewidth]{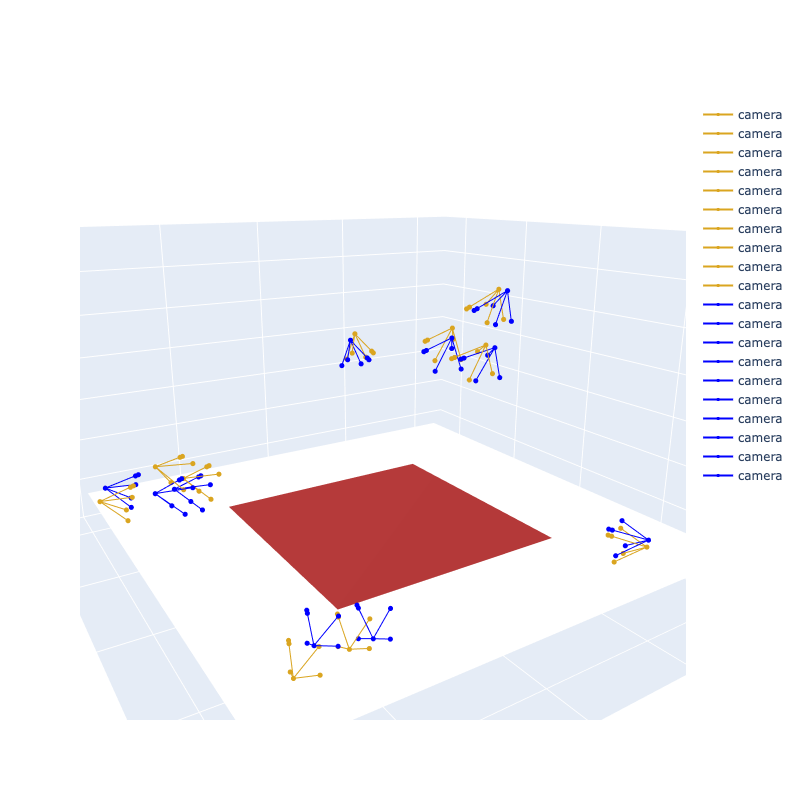}
\caption{
    \textbf{Camera noise} --
    Visualization of the effect of camera noise for $\sigma=0.1$, see \cref{sec:exp:sensitivity}.
    \emph{Blue}: original cameras,
    \emph{brown}: cameras with noise,
    \emph{red}: scene bounding box containing all objects.
\vspace{-4mm}
}
\label{fig:app:camnoise}
\end{figure}

%% file: figs/appendix_multishapenet.tex
\newcommand\amspic[1]{
\includegraphics[width=0.074\linewidth]{imgs/ams/#1}
}

\begin{figure*}[t]
    \setlength{\tabcolsep}{0mm}
    \def\arraystretch{1}
    \begin{tabular}{ccccccccccccc}
        \multicolumn{5}{c}{Input Views} & Target & LFN~\cite{sitzmann2021lfn} & PN~\cite{Yu21cvpr_pixelNeRF} & \lit{} & \uplit{} & 1-\lit{} & \vlit{} & Depth \\
        \cmidrule(lr){1-5}\cmidrule(lr){6-6}\cmidrule(lr){7-7}\cmidrule(lr){8-8}\cmidrule(lr){9-9}\cmidrule(lr){10-10}\cmidrule(lr){11-11}\cmidrule(lr){12-12}\cmidrule(lr){13-13}
        \amspic{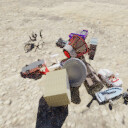} & \amspic{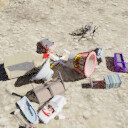} & \amspic{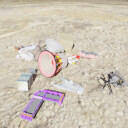} & \amspic{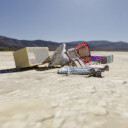} & \amspic{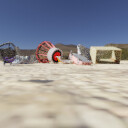} & \amspic{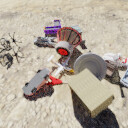} & \amspic{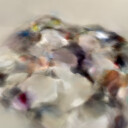} & \amspic{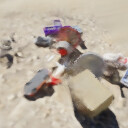} & \amspic{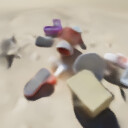} & \amspic{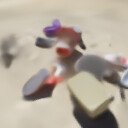} & \amspic{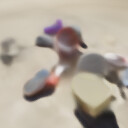} & \amspic{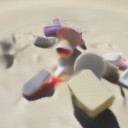} & \amspic{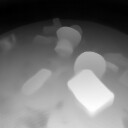} \\
        \amspic{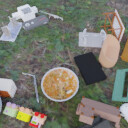} & \amspic{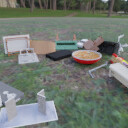} & \amspic{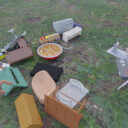} & \amspic{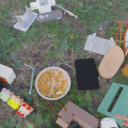} & \amspic{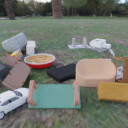} & \amspic{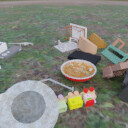} & \amspic{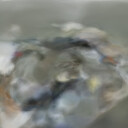} & \amspic{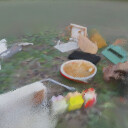} & \amspic{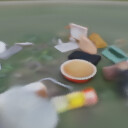} & \amspic{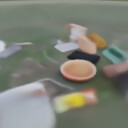} & \amspic{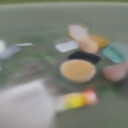} & \amspic{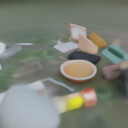} & \amspic{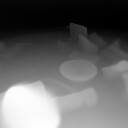} \\
        \amspic{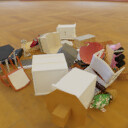} & \amspic{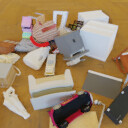} & \amspic{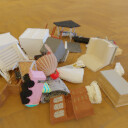} & \amspic{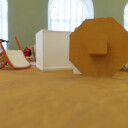} & \amspic{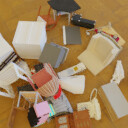} & \amspic{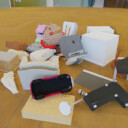} & \amspic{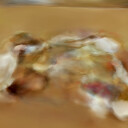} & \amspic{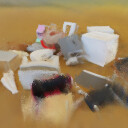} & \amspic{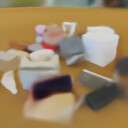} & \amspic{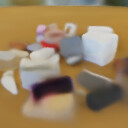} & \amspic{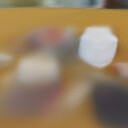} & \amspic{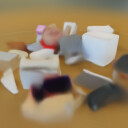} & \amspic{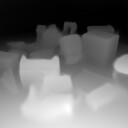} \\
        \amspic{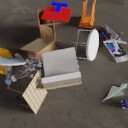} & \amspic{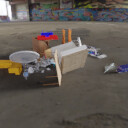} & \amspic{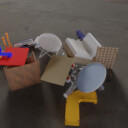} & \amspic{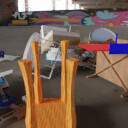} & \amspic{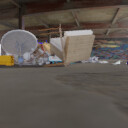} & \amspic{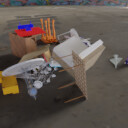} & \amspic{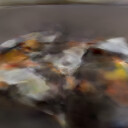} & \amspic{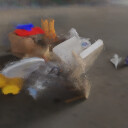} & \amspic{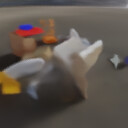} & \amspic{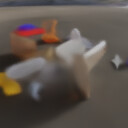} & \amspic{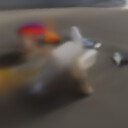} & \amspic{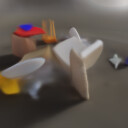} & \amspic{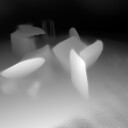} \\
        \amspic{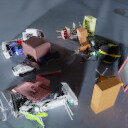} & \amspic{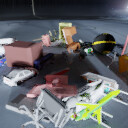} & \amspic{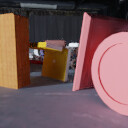} & \amspic{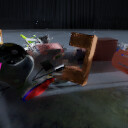} & \amspic{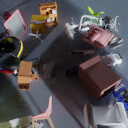} & \amspic{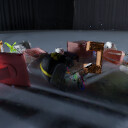} & \amspic{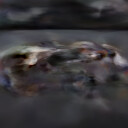} & \amspic{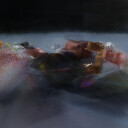} & \amspic{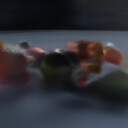} & \amspic{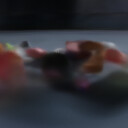} & \amspic{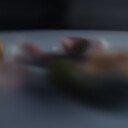} & \amspic{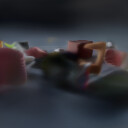} & \amspic{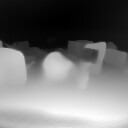} \\
        \amspic{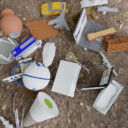} & \amspic{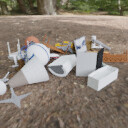} & \amspic{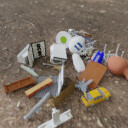} & \amspic{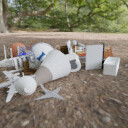} & \amspic{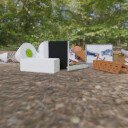} & \amspic{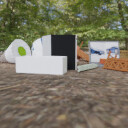} & \amspic{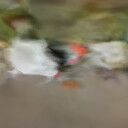} & \amspic{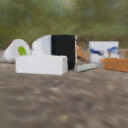} & \amspic{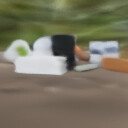} & \amspic{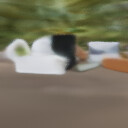} & \amspic{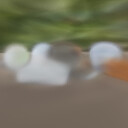} & \amspic{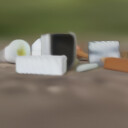} & \amspic{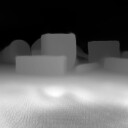} \\
        \amspic{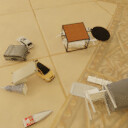} & \amspic{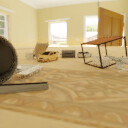} & \amspic{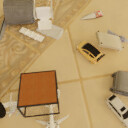} & \amspic{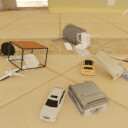} & \amspic{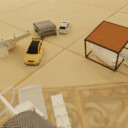} & \amspic{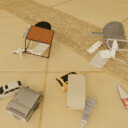} & \amspic{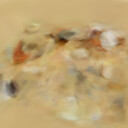} & \amspic{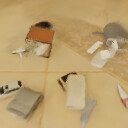} & \amspic{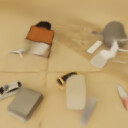} & \amspic{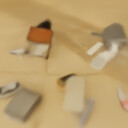} & \amspic{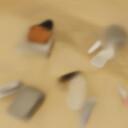} & \amspic{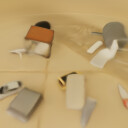} & \amspic{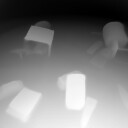} \\
        \amspic{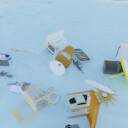} & \amspic{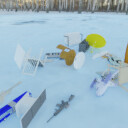} & \amspic{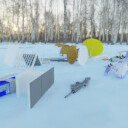} & \amspic{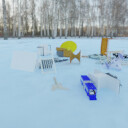} & \amspic{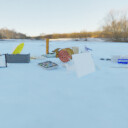} & \amspic{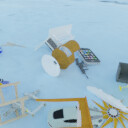} & \amspic{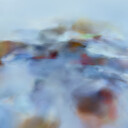} & \amspic{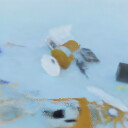} & \amspic{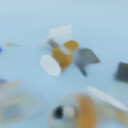} & \amspic{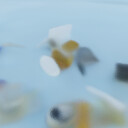} & \amspic{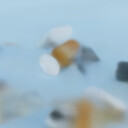} & \amspic{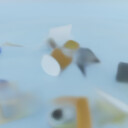} & \amspic{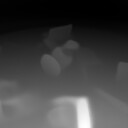} \\
        \amspic{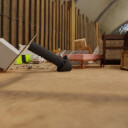} & \amspic{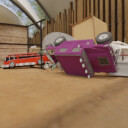} & \amspic{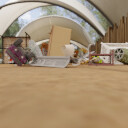} & \amspic{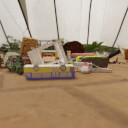} & \amspic{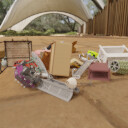} & \amspic{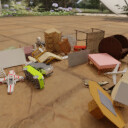} & \amspic{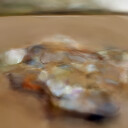} & \amspic{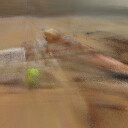} & \amspic{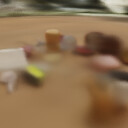} & \amspic{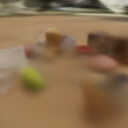} & \amspic{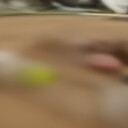} & \amspic{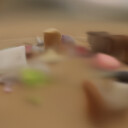} & \amspic{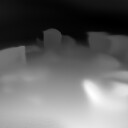} \\
        \amspic{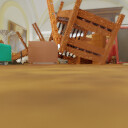} & \amspic{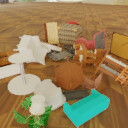} & \amspic{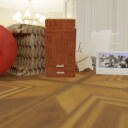} & \amspic{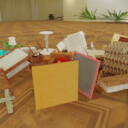} & \amspic{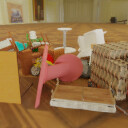} & \amspic{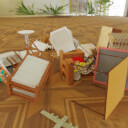} & \amspic{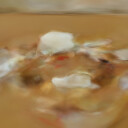} & \amspic{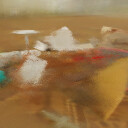} & \amspic{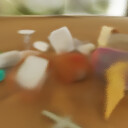} & \amspic{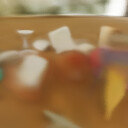} & \amspic{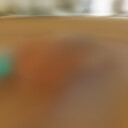} & \amspic{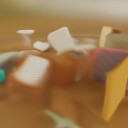} & \amspic{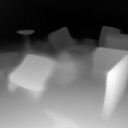} \\
        \amspic{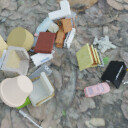} & \amspic{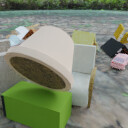} & \amspic{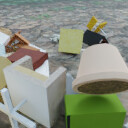} & \amspic{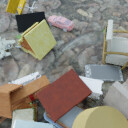} & \amspic{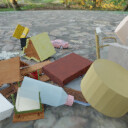} & \amspic{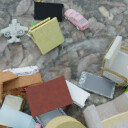} & \amspic{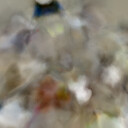} & \amspic{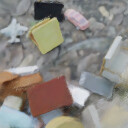} & \amspic{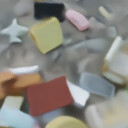} & \amspic{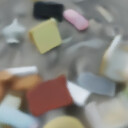} & \amspic{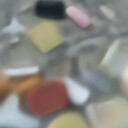} & \amspic{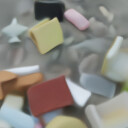} & \amspic{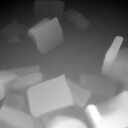} \\
        \amspic{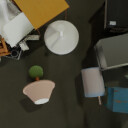} & \amspic{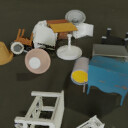} & \amspic{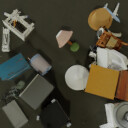} & \amspic{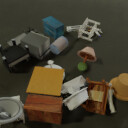} & \amspic{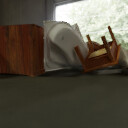} & \amspic{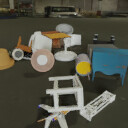} & \amspic{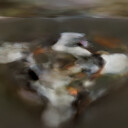} & \amspic{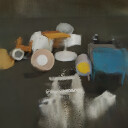} & \amspic{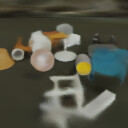} & \amspic{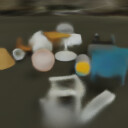} & \amspic{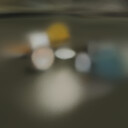} & \amspic{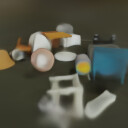} & \amspic{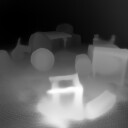} \\
        \amspic{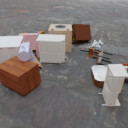} & \amspic{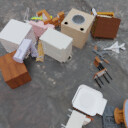} & \amspic{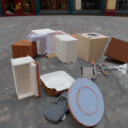} & \amspic{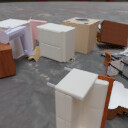} & \amspic{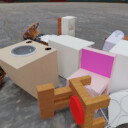} & \amspic{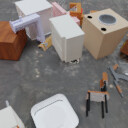} & \amspic{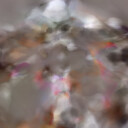} & \amspic{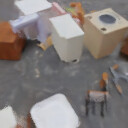} & \amspic{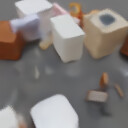} & \amspic{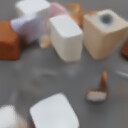} & \amspic{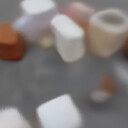} & \amspic{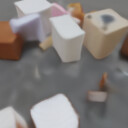} & \amspic{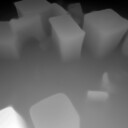} \\
        \amspic{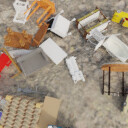} & \amspic{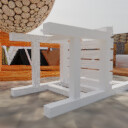} & \amspic{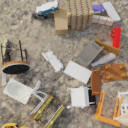} & \amspic{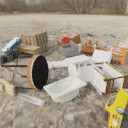} & \amspic{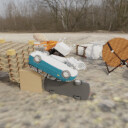} & \amspic{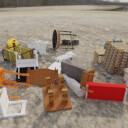} & \amspic{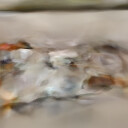} & \amspic{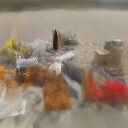} & \amspic{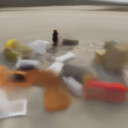} & \amspic{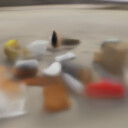} & \amspic{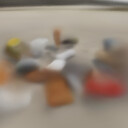} & \amspic{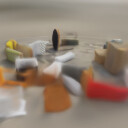} & \amspic{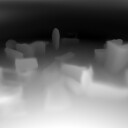} \\
    \end{tabular}
    \caption{
        \textbf{Qualitative results on MultiShapeNet} --
        More results on the \msn{} dataset from LFN~\cite{sitzmann2021lfn}, PixelNeRF~\cite{Yu21cvpr_pixelNeRF}, and \lit{} including variations with unposed inputs (\uplit{}), only the leftmost input image (1-\lit{}), and volumetric rendering (\vlit{} \& depth).
    } 
    \label{fig:app:multishapenet}
\end{figure*}